\documentclass[letterpaper]{article} % DO NOT CHANGE THIS
\usepackage{aaai23}  % DO NOT CHANGE THIS
\usepackage{times}  % DO NOT CHANGE THIS
\usepackage{helvet}  % DO NOT CHANGE THIS
\usepackage{courier}  % DO NOT CHANGE THIS
\usepackage[hyphens]{url}  % DO NOT CHANGE THIS
\usepackage{graphicx} % DO NOT CHANGE THIS
\usepackage{natbib}  % DO NOT CHANGE THIS AND DO NOT ADD ANY OPTIONS TO IT
\usepackage{caption} % DO NOT CHANGE THIS AND DO NOT ADD ANY OPTIONS TO IT
\usepackage{amsmath,amssymb} % define this before the line numbering.
\usepackage{amsmath}
\usepackage{colortbl}
\usepackage{bm}
\usepackage{times}
\usepackage{epsfig}
\newcommand{\tabincell}[2]{\begin{tabular}{@{}#1@{}}#2\end{tabular}}
 \usepackage{transparent}
\usepackage{mathrsfs}
\usepackage{dashrule}

\usepackage{multirow}
\usepackage{bbding}
\usepackage[flushleft]{threeparttable}
\usepackage{caption}
\usepackage{subcaption}
\usepackage{tabularx}
\usepackage{placeins}
\usepackage{booktabs}

\usepackage[ruled,linesnumbered,boxed]{algorithm2e}
\usepackage{bm}
\definecolor{MyGray}{gray}{0.9}
\makeatletter
\newcommand{\thickhline}{%
    \noalign {\ifnum 0=`}\fi \hrule height 1pt
    \futurelet \reserved@a \@xhline
}

\newcommand*\rfrac[2]{{}^{#1}\!/_{#2}}

\usepackage{xspace}

\makeatletter
\DeclareRobustCommand\onedot{\futurelet\@let@token\@onedot}
\def\@onedot{\ifx\@let@token.\else.\null\fi\xspace}

\def\eg{\emph{e.g}\onedot} 
\def\ie{\emph{i.e}\onedot} 
 
 \def\vs{\emph{vs}\onedot}
\def\wrt{w.r.t\onedot} 
\def\etal{\emph{et al}\onedot}
\makeatother

\title{Dynamic Appearance: A Video Representation\\ 
for Action Recognition with Joint Training}
\author {
    Guoxi Huang and Adrian G. Bors
}

\affiliations{
    Department of Computer Science, University of York, York YO10 5GH, UK\\
    gh825@york.ac.uk, adrian.bors@york.ac.uk
}

\begin{document}

\maketitle

\begin{abstract}
Static appearance of video may impede the ability of a deep neural network to learn motion-relevant features in video action recognition. In this paper, we introduce a new concept, Dynamic Appearance (DA), summarizing the appearance information relating to movement in a video while filtering out the static information considered unrelated to motion.
We consider distilling the dynamic appearance from raw video data as a means of efficient video understanding. To this end, we propose the Pixel-Wise Temporal Projection (PWTP), which projects the static appearance of a video into a subspace within its original vector space, while the dynamic appearance is encoded in the projection residual describing a special motion pattern.
Moreover, we integrate the PWTP module with a CNN or Transformer into an end-to-end training framework, which is optimized by utilizing multi-objective optimization algorithms. 
We provide extensive experimental results on four action recognition benchmarks: Kinetics400, Something-Something V1, UCF101 and HMDB51.
\end{abstract}

\section{Introduction}
Video data is collected across both space and time, is high-dimensional and contains substantial redundancy. This is challenging to any model aiming to differentiate between relevant and irrelevant features.
% Video data is represented by a huge amount of data representing multi-dimensional spatio-temporal information. Video data also contains substantial redundancies. These circumstances are challenging to any model aiming to differentiate between relevant and irrelevant spatio-temporal features.
% Although many spatio-temporal neural networks such as I3D~\cite{carreira2017quo} have been proposed for action recognition in videos,  learning spatio-temporal feature is still challenging.
% When Convolutional Neural Network (CNN) has become the dominant solution in various visual applications by virtue of its excellent feature learning ability, yet spatio-temporal feature learning is still challenging. 
While leading-edge video architectures~\cite{carreira2017quo,wang2018non,feichtenhofer2019slowfast,feichtenhofer2020x3d,xie2018rethinking,bertasius2021space,fan2021multiscale} have been proposed for learning spatio-temporal features from raw video data, a quintessential spatio-temporal representation remains elusive.

\begin{figure*}[!t]
\begin{center}
\includegraphics[width=.88\textwidth]{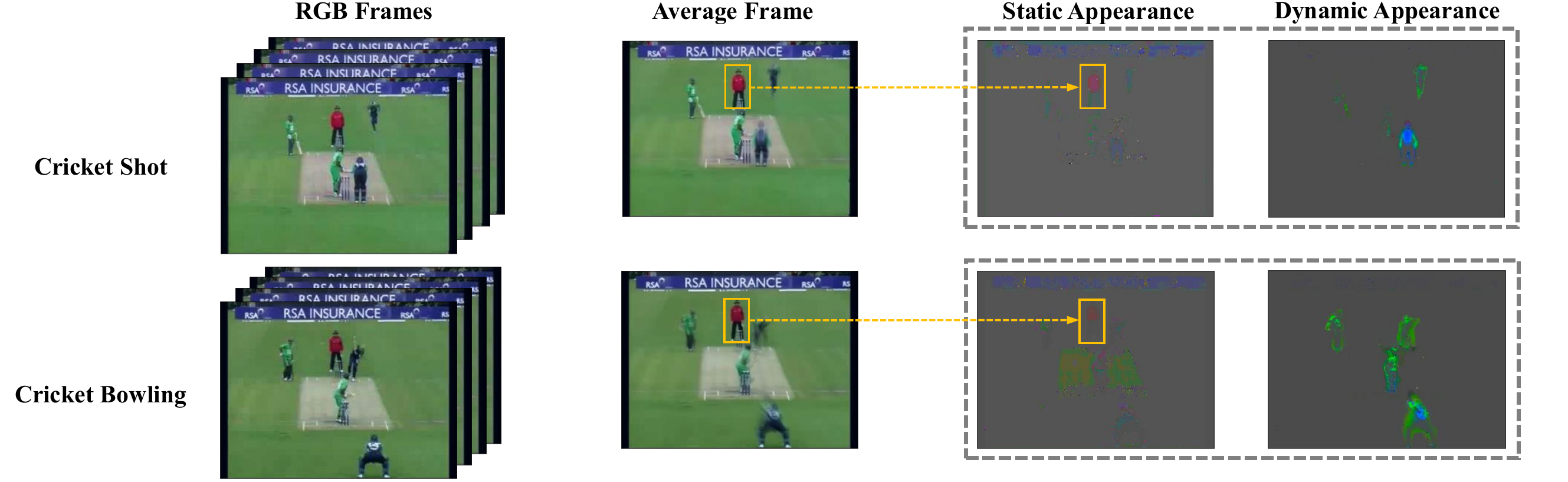}
\end{center}
%%%%\vspace*{-0.4cm}
\caption{Different actions recorded within the same environment have very similar static appearances, which could cause confusion in classifiers. Using the Dynamic Appearance as an input resource would avoid such confusions, as it only contains the visual information related to the movement. For instance, the person in the bounding box does not move and is not included in the dynamic appearances but rather in the static appearances.}
\label{fig: rgb_avg_da}
%%%\vspace*{-0.4cm}
\end{figure*}

The idea of characterizing movement in a video with appearance or silhouette was established early in traditional appearance-based temporal matching methods~\cite{ahad2008motion,bradski2002motion}. 
Drawing upon this, we propose a novel video representation, termed Dynamic Appearance, which summarizes the appearance information relating to movement in a video sequence while filtering out the static information considered unrelated to motion. The filtered static information that does not change from frame to frame is temporarily referred to as the Static Appearance. To better understand the concept, we depict the dynamic appearance and static appearance of a video in Fig.~\ref{fig: rgb_avg_da}. 

Our motivation for this study stems from the observation that many videos labeled with different action categories are shot in the same environment. Taking the videos shown in Fig.~\ref{fig: rgb_avg_da} as an illustrative example, the two videos have different action labels (`Cricket Shot' \vs `Cricket Bowling') but they are actually two short video clips shot in the same environment.
In this case, the static appearances and scene information of these videos are very similar, describing the same object and background, which hinders deep networks that take RGB frames as input from learning relevant spatio-temporal features.
In the two-stream architecture~\cite{simonyan2014two}, separately modeling appearance and motion by adding additional optical flow input modality could avoid the distraction caused by static appearance information. However, the computational complexity of two-stream-based architecture~\cite{simonyan2014two,feichtenhofer2016spatiotemporal,feichtenhofer2016convolutional,wang2016temporal,carreira2017quo} is double that of the single-stream version. We demonstrate that a network that takes dynamic appearance as input can achieve accuracy on par with the two-stream networks while requiring less than half the computation of two-stream networks.

To disentangle the dynamic appearance from the static appearance in a video, we propose the Pixel-Wise Temporal Projection (PWTP),  a lightweight and backbone-agnostic module, which is used to project the temporal information into a subspace of its original temporal dimension. The resulting projection encodes the video's static appearance while the projection residual encodes the dynamic appearance.
The dynamic appearance learning enabled by PWTP can firstly be regarded as an unsupervised learning method, trained on unlabeled video data. Alternatively, we can integrate the PWTP module with a network into a unified framework, which is
optimized with joint training algorithms. As a result, the framework would produce the dynamic appearance on the fly while performing action recognition. The dynamic appearance learning works as an auxiliary task to the primary recognition task.
It is difficult for a network to learn action-relevant features with the recognition criteria because of the impediment of irrelevant static appearance features. However, doing this would be easier for the dynamic appearance learning task. 
By jointly training the two tasks, the model can better generalize on the action recognition task, as demonstrated in our experiments.
We perform extensive experiments on four standard video benchmarks: Kinetics400~\cite{carreira2017quo}, Something-Something V1~\cite{goyal2017something}, UCF101~\cite{soomro2012ucf101} and HMDB51~\cite{kuehne2011hmdb}, where we show that PWTP greatly
improves the performance in terms of accuracy. The experimental results successfully demonstrate that using the dynamic appearance extracted by PWTP can enable better feature learning in networks.

\section{Related Work}

\subsubsection{Spatio-temporal Networks}
With the tremendous success of Convolutional Neural Networks(CNNs) on image classification tasks~\cite{krizhevsky2012imagenet,simonyan2014very,szegedy2015going,szegedy2016rethinking}, various CNN-based architectures~\cite{karpathy2014large,yue2015beyond,wang2016temporal,zhu2018hidden,feichtenhofer2016convolutional,wang2017spatiotemporal,wang2021tdn} have been proposed for spatio-temporal modeling in video recognition.
Representatively, the two-stream architecture~\cite{simonyan2014two} and its variants~\cite{wang2016temporal,carreira2017quo,bilen2017action,fan2018end,zhu2018hidden,lin2019tsm} separately model appearance and motion by taking RGB video frames and the optical flow as inputs. 
Some published works~\cite{taylor2010convolutional,tran2015learning,carreira2017quo,hara2018can} employ the computationally intensive 3D convolution to their models. Several innovative and efficient spatio-temporal networks such as P3D \cite{qiu2017learning}, R(2+1)D \cite{tran2018closer}, CSN \cite{tran2019video} and X3D \cite{feichtenhofer2020x3d} are proposed to seek a trade-off between speed and accuracy.
With the emergence of Vision Transformer (ViT)~\cite{dosovitskiy2021image}, the backbone architectures in video recognition are shifted from CNNs to Transformers in recent work~\cite{dosovitskiy2021image,bertasius2021space,fan2021multiscale,arnab2021vivit}.

\subsubsection{Motion Representation}
Optical flow as a short-term motion representation has been widely used in action recognition. Flow Net series~\cite{ilg2017flownet,dosovitskiy2015flownet} and AutoFlow~\cite{sun2021autoflow} improves the quality of optical flow estimation by using deep learning. Some works~\cite{ng2018actionflownet,zhu2018hidden} attempt to integrate optical flow estimation and the action recognition model into an end-to-end training framework.
However, optical flow represents motion with instantaneous image velocities, which loses the appearance information, and therefore could cause inaccurate classification of videos in different scenes.  
Derived from the optical flow, OFF~\cite{sun2018optical}, Flow-of-Flow~\cite{piergiovanni2019representation} and other flow-based methods are proposed for fast motion feature learning. 
The seminal work~\cite{bilen2017action} utilizes the rank pooling~\cite{fernando2015modeling} or approximate rank pooling~\cite{bilen2017action} machine to summarize both the static and dynamic visual information of a video into a few RGB images, called dynamic images. It is worth noting that vanilla PCA cannot achieve the same result as rank pooling~\cite{fernando2015modeling} because PCA is a generic orthogonalization method for dimension reduction by selecting a set of orthogonal vectors, the output of which would not be adapted to the 2D or 3D layout. SVM pooling~\cite{wang2018video} generates a video representation similar to dynamic images by exploiting the boundary information of SVM. 
Nevertheless, dynamic images result in poor discrimination between the moving objects and background, which harms performance when recorded videos are affected by camera shaking.
Different to the existing methods, the proposed PWTP achieves the disentanglement of video's static and dynamic appearance. The dynamic appearance can be a preferable input resource to networks, encoding the temporal motion information as well as the visual information considered related to movement.

\subsubsection{Joint Training}
Joint Training, one of the guises of multi-task learning, can be regarded as a form of inductive transfer, helping a model generalize better by introducing an inductive bias. Ng~\etal~\cite{ng2018actionflownet} proposed ActionFlowNet for jointly estimating optical flow and recognizing actions.
Similarly, TVNet~\cite{fan2018end} jointly trained the video representation learning and action recognition tasks.
In our case, the representation learning for the static and dynamic appearance disentanglement of video is an auxiliary task to the primary action recognition task. This would allow the model to eavesdrop, \ie learn the motion-relevant representation through the dynamic appearance learning task. Existing methods balance the recognition loss and the objective functions of their representation learning with some fixed scale parameters. In our study, in order to stabilize the feature learning process we analyze various joint training scenarios, such as the multiple gradient descent algorithm~\cite{desideri2012multiple}.

\section{Methodology}
\subsection{Pixel-Wise Temporal Projection}
\label{sec:Pixel-Wise Temporal Projection}

In order to achieve video static and dynamic appearance disentanglement, we propose the Pixel-Wise Temporal Projection (PWTP) operator. 
Given a $T$ frame video clip $\mathbf{x}$ of spatial resolution $H \times W$, we flatten it along the spatial dimensions, so $\mathbf{x}=\{ \mathbf{x}_i \}^{HW}_{i=1}$, $\mathbf{x}_i \in \mathbb{R}^{T \times C}$, and $C$ is the number of channels.

PWTP is capable of summarizing the video information from $T$ frames into $D$ frames, and $D << T$. 
Firstly, for every position $i$ in the spatial dimensions, PWTP learns a particular matrix \(\mathbf{A}_i \in \mathbb{R}^{T \times D}\), the column space of which defines a subspace (\ie hyperplane) in $\mathbb{R}^T$. The columns of matrix $\mathbf{A}_i$ are linearly independent. Then the PWTP projects $\mathbf{x}_i$ onto the corresponding column space of $\mathbf{A}_i$, resulting $\widehat{\mathbf{x}}_i$, given by
\begin{equation} 
\label{eq_projection}
\begin{aligned}
&\mathbf{y}_i = (\mathbf{A}_i^\intercal \mathbf{A}_i)^{-1} \mathbf{A}_i^\intercal \mathbf{x}_i, \\
&\widehat{\mathbf{x}}_i = \mathbf{A}_i \mathbf{y}_i, \quad  i=1,\ldots,HW,
\end{aligned}
\end{equation}
where $\mathbf{y}=({\bf y}_1  \ldots {\bf y}_{HW})^T$, $\mathbf{y}_i \in \mathbb{R}^{D \times C} $ are the mapping coefficients. 
Eq.~\eqref{eq_projection} represents a standard projection process in the temporal dimension, which is repeatedly applied to all spatial positions. 
The main component of a video can be considered to be made up of the video's static information, as the static information repeatedly appears in many frames, showing high redundancy.
When the dimension of the subspace (\ie rank($\mathbf{A}_i)$) is much smaller than the original temporal dimension $T$ in a video, its information storage capacity is significantly reduced, such that the projection output $\widehat{\mathbf{x}}$ tends to encode the repeated information, characterizing static appearance that does not change from frame to frame.

The production of input-specific $\mathbf{A}=( \mathbf{A}_1 \ldots \mathbf{A}_{HW} )$ is realized in two steps: 1) producing a temporal relation descriptor for every spatial position $i$; 2) a Multilayer Perceptron (MLP) takes as input the temporal relation descriptors and outputs the tensor $\mathbf{A}$. In the first step for the temporal relation description, we pass the input $\mathbf{x}$ to a spatial convolution $\mathcal{F}$ with the kernel size of $k \times k$ and stride of $s$ for local spatial information aggregation:  
\begin{equation} 
\label{eq_conv1}
\begin{aligned}
&\tilde{\mathbf{x}} = \mathcal{F}^{k \times k}_{s}(\mathbf{x}; \theta),
\end{aligned}
\end{equation}
where $\theta$, of size $k \times k \times C \times C'$, denotes the kernel weights.
The spatial convolution $\mathcal{F}$ maps the tensor $\mathbf{x}$ to $\tilde{\mathbf{x}}$ of size ${\frac{HW}{s^2}\times T \times C'}$, which significantly reduces the computational cost for PWTP when setting the stride $s > 1$. This is a non-trivial trick that allows the pixel-wise projection to be implemented within a feasible computation budget.
Subsequently, we have $\tilde{\mathbf{x}}=\{\tilde{\mathbf{x}}_i\}^{HW/s^2}_{i=1}$ and $\tilde{\mathbf{x}}_i = \left \{ \tilde{\mathbf{x}}^t_i \right \} ^T_{t=1}$, where $\tilde{\mathbf{x}}^t_i \in \mathbb{R}^C$.
Then, the temporal relation descriptor $\mathbf{u}_i$ for every spatial position $i$ is obtained as 
\begin{equation} 
\label{eq_descriptor}
\begin{aligned}
\mathbf{u}_i = \left \{ \rfrac{1}{C'} \tilde{\mathbf{x}}^{t_1\intercal}_i\tilde{\mathbf{x}}^{t_2}_i \; \big{|} \; t_1\neq t_2 ,  1 \leqslant  t_1,t_2 \leqslant  T \right \}.
\end{aligned}
\end{equation}

Consequently, $\mathbf{u}_i$ is represented as a vector $\mathbf{u}_i \in \mathbb{R}^{\frac{T \times (T-1)}{2}}$ with ascending indices of $t_1$, $t_2$, and $\mathbf{u} = [\mathbf{u}_1, \ldots, \mathbf{u}_{\frac{HW}{s^2}}]$.
As each element in $\mathbf{u}_i$ describes the similarity between a pair of frames, such that $\mathbf{u}_i$ can represent the relationship between any two frames in the video clip. By taking $\mathbf{u}$ as the input to an encoder, the projection machine would have a better understanding of the temporal relationships in the video, and so provide a better temporal projection mechanism.

The convolution $\mathcal{F}^{k \times k}_s (\cdot; \theta)$ from Eq.~\eqref{eq_conv1} is important for implementing temporal projection at pixel-level, as it encodes the local spatial information of each position $i$ into the local descriptor $\mathbf{u}_i$, which leads to the projection result $\widehat{\mathbf{x}}$ representing smooth textures. Without Eq.~\eqref{eq_conv1}, the projection in each spatial position would be performed completely independently, which would fail to consider that the values of neighboring pixels are continuous, resulting in unsmooth spatial textures.

In the second step of the generation of $\mathbf{A}$, we utilize an MLP to process the temporal relation descriptors $\mathbf{u}$ and generate $\mathbf{A}$:
\begin{equation} 
\label{eq:mlp_resahpe}
\begin{aligned}
& \mathbf{A} = Upsample(\rm{MLP}(\mathbf{u})), \\
& Reshape: \mathbf{A} \in \mathbb{R}^{HW \times TD} \rightarrow  \mathbf{A} \in \mathbb{R}^{HW \times T \times D}, \\
\end{aligned}
\end{equation}
where the 2D bilinear upsampling is employed in the spatial dimensions if convolution $\mathcal{F}$'s stride $s > 1$. Tensor reshaping is then performed at the end.  The linear independence of the columns of $\mathbf{A}_i$ can be implemented by properly initializing the MLP (\eg by avoiding to initialize all weights to zero).
The MLP in the PWTP module is made up of multiple fully-connected layers with a bottleneck design where the number of features is first reduced by a factor and then recovered to its original size. The residual connection~\cite{he2016deep} is applied after every bottleneck. Each fully-connected layer, except for the last one, is followed by a GELU non-linearity. More details about the MLP configurations are provided in Appendix~C from the Supplementary Material (SM).

\subsection{Projection Residual as Dynamic Appearance}

The projection residual in spatial position $i$ is given by
\begin{equation} 
\label{eq_proj_res}
\begin{aligned}
\mathbf{p}_i = \mathbf{x}_i - \widehat{\mathbf{x}}_i.
\end{aligned}
\end{equation}

Note that when the dimension $D$ of the subspace (\ie rank($\mathbf{A}_i)$) is much smaller than the original temporal dimension $T$ in a video, the projection $\widehat{\mathbf{x}}_i$ is forced to encode the video's static information, because static information repeatedly appear in consecutive frames, construing the main components of $\mathbf{x}_i$. Subtracting the common static features by the original $\mathbf{x}_i$ will generate the dynamic appearance (DA).
The projection residual $\mathbf{p}$ preserves the pure dynamic appearance information of the video and filters out the static appearance. Considering that many spatial positions in the projection residual do not carry visual information after optimization, $\mathbf{p}_i \approx \mathbf{0}$, we average $\mathbf{p}$ along the temporal direction to reduce the tensor size:
\begin{equation} 
\label{eq_mri}
\begin{aligned}
\bar{\mathbf{p}}_i = \frac{1}{T}\sum^{T}_{t=1}\mathbf{p}^t_i.
\end{aligned}
\end{equation}

\subsection{Action Recognition Framework}
\label{sec: Action Recognition Pipeline}

\begin{figure}[!t]
\begin{center}
\includegraphics[width=.485\textwidth]{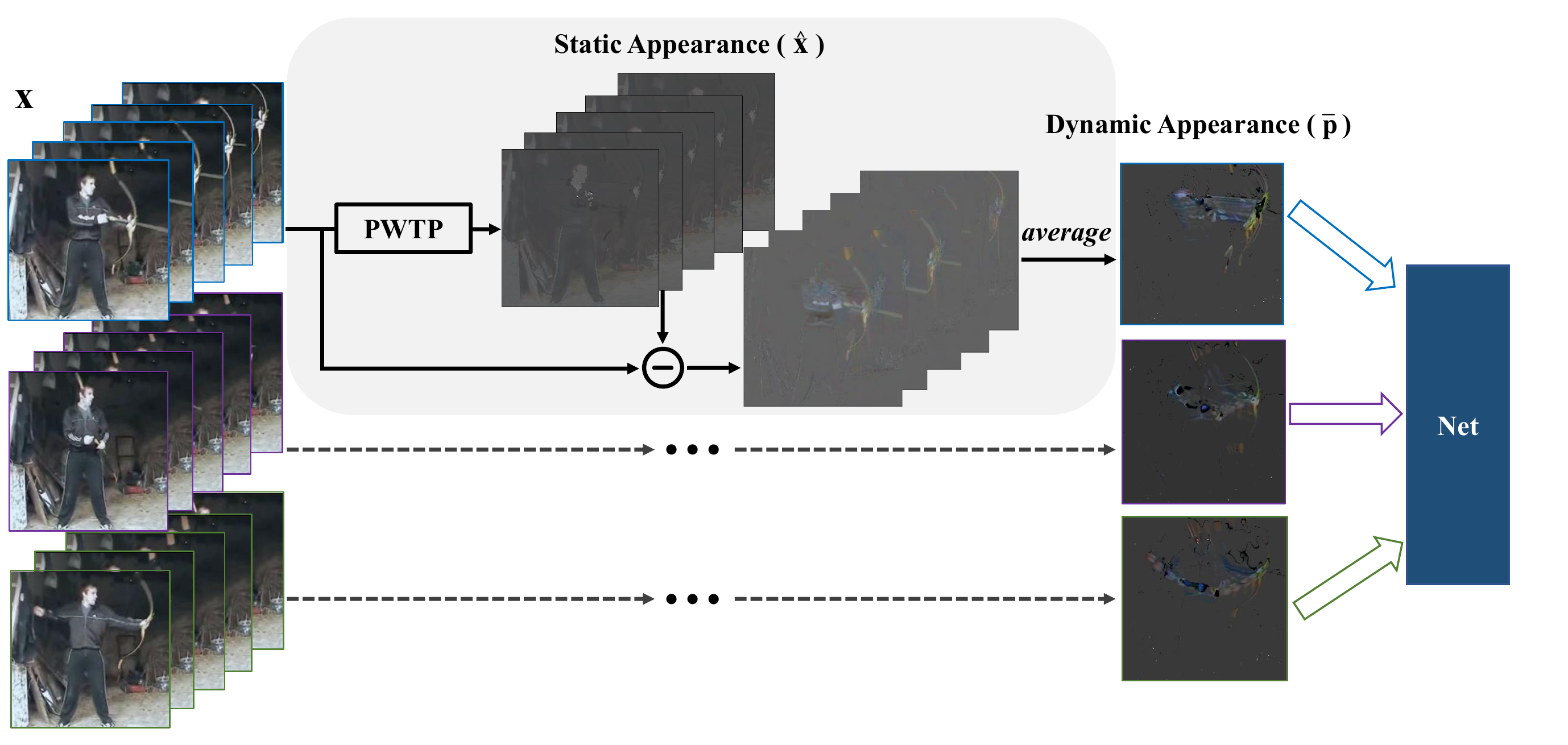}
\end{center}
%%%\vspace{-0.3cm}
\caption{The video recognition framework enabled by PWTP, in which PWTP separates the dynamic appearance from the static appearance. Subsequently, the shared backbone network feeds on the dynamic appearances.}
\label{fig: arch}
%%%\vspace{-0.5cm}
\end{figure}

\noindent The recognition framework that unifies the PWTP module and a backbone network is presented in Fig.~\ref{fig: arch}. We sample and then divide a video along the temporal dimension into $S$ segments, with each segment containing $T$ consecutive video frames. The proposed PWTP module is applied on every segment. As a result, the $S \times T$ RGB frames are projected into $S$ frames of dynamic appearance. 

By taking the dynamic appearance $\bar{\mathbf{p}}$ as input, a network can avoid the distraction caused by the static appearance ($\widehat{\mathbf{x}}$) when discriminating the actions in some similar scenes, and therefore can focus on learning high-level motion-relevant features. 
In order to evaluate the generalization ability of our method we adapt it to different backbone architectures, including TSM R50~\cite{lin2019tsm} and X3D-XS, X3D-M~~\cite{feichtenhofer2020x3d} and ViT-B~\cite{dosovitskiy2021image}. 
In order to adapt ViT-B to video applications, we duplicate the positional embedding for each segment to encode the spatio-temporal positional information. Instead of using the class token from the last layer of ViT for prediction, we average the entire sequence of embeddings from the last layer to produce the prediction. Dynamic appearance is shown experimentally to steadily improve the performance of different networks.

\subsection{Joint Training}
\label{sec:joined_training}
The optimization objective of PWTP is defined to minimize the Euclidean Norm of Projection Residual (ENoPR): 
\begin{equation}
\label{eq:loss_proj}
\begin{aligned}
\mathcal{L}^{1} = \frac{1}{HWC} \sum^{HWC}_{j=1} \parallel \mathbf{\tilde{p}}_j \parallel^2,
\end{aligned}
\end{equation}
where $\mathbf{\tilde{p}} \in \mathbb{R}^{HWC \times T}$ denotes the flattened version of $\mathbf{p}$ along the space and channels. $\mathcal{L}^{1}$ can be defined as a loss function $\mathcal{L}^{1}(\Theta^1)$ \wrt PWTP's weight parameters $\Theta^1$. For the case of $N$ data points, the ENoPR is expressed by:
\begin{equation}
\label{eq:loss_proj_n_point}
\begin{aligned}
\widehat{\mathcal{L}}^1(\Theta^1) = \frac{1}{N} \sum^N_{n=1} \mathcal{L}^{1}_{n}(\Theta^1).
\end{aligned}
\end{equation}

% We want to build an end-to-end training pipeline, which minimizes the projection residual of the PWTP and the cross-entropy loss of the backbone network. 

% As seen in Figure~\ref{fig: dynamic_still_seprate}, the PWTP, optimized independently, tends to clearly filter out the static information such as background, which is essential for scene perception. Some weak static details is expected to exist in the MRI for the best performance.
% This can be achieved by joint training of the backbone network and PWTP. 
The loss function of action recognition is denoted by $\widehat{\mathcal{L}}^{2}(\Theta^2)$ in independent training, $\widehat{\mathcal{L}}^{2}(\Theta^1,\Theta^2)$ in joint training, where PWTP's parameters $\Theta^1$ are shared between the two tasks.
The joint training of $\widehat{\mathcal{L}}^1(\Theta^1)$ and $\widehat{\mathcal{L}}^2(\Theta^1,\Theta^2)$ forms the multi-objective optimization~: 
\begin{equation}
\label{eq:MTL}
\begin{aligned}
\min_{\Theta^1, \Theta^2} \quad \left[ \alpha \widehat{\mathcal{L}}^{1}(\Theta^1) + (1-\alpha) \widehat{\mathcal{L}}^{2}(\Theta^1,\Theta^2) \right], \quad0 \leqslant \alpha \leqslant 1,
\end{aligned}
\end{equation}
where the weight $\alpha$, used to scale the contribution of the two loss functions, can be fixed, as in ~\cite{fan2018end}, or can be adaptive.

\subsection{Multiple Gradient Descent Algorithm}
\label{sec:mgda}
For the adaptive scale $\alpha$, we adopt the Multiple Gradient Descent Algorithm (MGDA)~\cite{desideri2012multiple,sener2018multi}, in which the goal of multi-objective optimization is to achieve Pareto-optimality. A solution is said to be of Pareto-stationary if and only if there is $0 \leqslant \alpha \leqslant 1$ such that:
\begin{equation}
\label{eq:pareto_sta}
\begin{aligned}
\bullet & \quad \alpha \nabla_{\Theta^1} \widehat{\mathcal{L}}^{1}(\Theta^1) + (1-\alpha) \nabla_{\Theta^1}\widehat{\mathcal{L}}^{2}(\Theta^1,\Theta^2) = 0, \\
\bullet & \quad \nabla_{\Theta^2}\widehat{\mathcal{L}}^{2}(\Theta^1,\Theta^2) = 0.
\end{aligned}
\end{equation}
Pareto-stationarity is a necessary condition for Pareto-optimality. From~\cite{desideri2012multiple,sener2018multi}, the decent directions to Pareto-stationarity would be provided by solving the following optimization problem:
\begin{equation}
\label{eq:pareto_sta_direction}
\begin{aligned}
\min_{\alpha} \:\: \parallel \alpha \nabla_{\Theta^1} \widehat{\mathcal{L}}^{1}(\Theta^1) + (1-\alpha) \nabla_{\Theta^1}\widehat{\mathcal{L}}^{2}(\Theta^1,\Theta^2) \parallel^2_2 ,\\
0 \leqslant \alpha \leqslant 1,
\end{aligned}
\end{equation}
which is a one-dimensional quadratic function of $\alpha$ with an analytical solution:
\begin{equation}
\label{eq:pareto_sta_solution}
\begin{aligned}
\alpha' &= \frac{\big(\nabla_{\Theta^1}\widehat{\mathcal{L}}^{2}(\Theta^1,\Theta^2) - \nabla_{\Theta^1}\widehat{\mathcal{L}}^{1}(\Theta^1)\big)^\intercal \nabla_{\Theta^1}\widehat{\mathcal{L}}^{2}(\Theta^1,\Theta^2)}{\parallel \nabla_{\Theta^1}\widehat{\mathcal{L}}^{1}(\Theta^1) - \nabla_{\Theta^1}\widehat{\mathcal{L}}^{2}(\Theta^1,\Theta^2) \parallel^2_2}, \\
\widehat{\alpha} &= \max(\min(\alpha', 1),0).
% {+,{1 \atop \intercal}}
\end{aligned}
\end{equation}
Suppose we optimize the objective with mini-batch gradient descent, then the gradient is updated with Algorithm~1: 
\begin{algorithm}
    \small
	\caption{Optimization of the Joint Training}
	\label{alg:mgda}
	\KwIn{Training tasks: $\widehat{\mathcal{L}}^{1}(\Theta^1)$, $\widehat{\mathcal{L}}^{2}(\Theta^1,\Theta^2)$; Learning rate: $\eta$;}
	\BlankLine
	Initialize $\Theta^1, \Theta^2$ randomly;
	
	\While{\textnormal{not converged}}{
	Compute gradient: 
	$\nabla_{\Theta^1}\widehat{\mathcal{L}}^{1}(\Theta^1)$ and
	$\nabla_{\Theta^1}\widehat{\mathcal{L}}^{2}(\Theta^1, \Theta^2)$;
	
	Update $\alpha$ by solving Eq.~\eqref{eq:pareto_sta_solution};
	
	$\Theta^1 = \Theta^1  - \eta \left( \alpha\nabla_{\Theta^1}\widehat{\mathcal{L}}^{1}(\Theta^1)+(1-\alpha)\nabla_{\Theta^1}\widehat{\mathcal{L}}^{2}(\Theta^1, \Theta^2 ) \right)$;
	
	$\Theta^2 = \Theta^2  - \eta \nabla_{\Theta^2}\widehat{\mathcal{L}}^{2}(\Theta^1,\Theta^2)$;
	}
\end{algorithm}

\section{Experiments}
\subsubsection{Datasets} We conduct experiments on four action recognition datasets, including Kinetics400 (K400)~\cite{carreira2017quo}, UCF101~\cite{soomro2012ucf101}, HMDB51~\cite{kuehne2011hmdb} and Something-Something (SS) V1~\cite{goyal2017something}.
In K400, UCF101 and HMDB51, temporal relation is less important than the RGB scene information. Most videos from these datasets can be accurately classified by considering only their RGB scene information. 
In SS V1, many action categories are symmetrical. Discriminating such actions in videos requires modeling temporal data relationships.  Mini-Kinetics (Mini-K)~\cite{xie2018rethinking} is sub-set of K400 with 200 categories. Some ablation studies are experimented on Mini-K for fast exploration.

\subsubsection{Implementation Details}
Unless specified otherwise, our models adopt the following settings: the embedded PWTP module is configured with $T=8$, $D=1$. The convolution $\mathcal{F}^{k \times k}_{s}(\cdot; \theta)$ is configured with the kernel size of $9\times9$, stride $s=8$ and $C'=24$ output channels . 
We jointly train the dynamic appearance learning and action recognition tasks with the Multiple Gradient Descent Algorithm described in Sec.~\ref{sec:joined_training}. 
For SS V1, UCF101 and HMDB51, we use the sparse sampling strategy in TSN~\cite{wang2016temporal}, where a video is evenly divided into $S$ segments. For each segment, we select $T$ consecutive video frames to form a clip of $S \times T$ frames.
For K400~\cite{carreira2017quo}, we perform dense sampling, selecting $S \times T$ consecutive frames. The dynamic appearance extraction is performed on every $T$ consecutive frames. The training and inference protocols are provided in Appendix~A from the SM.

\begin{table*}[!t]
     \centering
     \small
      \setlength{\tabcolsep}{0.4em}
      \begin{tabular}{l c c c c c c c c}
      \toprule
      \bfseries \multirow{2}*{Method} & \bfseries \multirow{2}*{Modality}  &\bfseries \multirow{2}*{pretrain} &\bfseries \multirow{2}*{Backbone} & \bfseries Frames & \bfseries \multirow{2}*{FLOPs} & \bfseries \multirow{2}*{\#Param.} & \bfseries Top-1   & \bfseries  Top-5 \\
      & & & & \bfseries $\times$views &  & & \bfseries (\%) & \bfseries (\%) \\
      \midrule

      TRN~\cite{zhou2018temporal} & RGB+Flow &  IN-1K & BNInc. & (8$+$40)$\times$1 & N/A & 36.6M& 42.0 & -\\
      \cline{2-2}
    ir-CSN~\cite{tran2019video} &  \multirow{4}*{RGB} & None & 3D R152 & 32$\times$10 &74G$\times$10 & 29.7M & 49.3 & - \\
    NL I3D~\cite{wang2018non}  &  & K400 & 3D R50 & 32$\times$2 & 168G$\times$2 & N/A & 44.4 & 76.0\\
    \cline{3-3}
    TEA~\cite{li2020tea}  &  & \multirow{2}*{IN-1K}& R50 & 16$\times$30 & 70G$\times$30 & 24.4M& 52.3 & 81.9\\
    TDN~\cite{wang2021tdn}  &  &  & R101 & 16$\times$1 & & 132G$\times$1 & 55.3 &  83.3\\
    \midrule
    \midrule
    ViT-B$\dagger$~\cite{dosovitskiy2021image} & RGB & \multirow{2}*{IN-21K} & - & 8$\times$3 & 269G$\times$3 & 86.3M& 45.7 & 74.3 \\
    \textbf{ViT-B$\dagger$ (Ours)} & DA$_{S=8,T=4}$ &  & - & 8$\times$3 & 270G$\times$3 & 86.3M& 46.4 & 75.8 \\
    \midrule
    TSM~\cite{lin2019tsm} & RGB & \multirow{4}*{IN-1K}  & \multirow{4}*{R50} & 8$\times$6 & 42.9G$\times$6 & 23.8M & 48.7 & 77.9\\
      TSM~\cite{lin2019tsm} & Flow &  &   & 40 $\times$6 & N/A & 48.6M&  39.5 & 70.5\\
      TSM~\cite{lin2019tsm} & RGB+Flow &  &   & (8$+$40)$\times$3 & N/A & 48.6M&  50.6 & 80.1\\
    \textbf{TSM (ours)} & DA$_{S=8,T=4}$ &   &  & 8$\times$6 & 43.1G$\times$6 & 23.8M & 50.1 & 78.6\\

    \midrule
    X3D-XS~\cite{feichtenhofer2020x3d} & RGB & \multirow{4}*{K400}  & - & 4$\times$6 &0.6G$\times$6 & 3.33M &  40.6 & 70.5 \\
    \textbf{X3D-XS (ours)}& DA$_{S=4,T=8}$  & & - & 4$\times$6 & 0.8G$\times$6 & 3.34M &  43.1 & 73.4 \\
    
    X3D-M~\cite{feichtenhofer2020x3d} &  RGB & & - & 16$\times$6 &6.3G$\times$6 & 3.33M &  52.0 & 81.0 \\

    \textbf{X3D-M (ours)} & DA$_{S=16,T=3}$ &  & - & 16$\times$6 & 7.0G$\times$6 & 3.34M & 53.7 & 82.1 \\
    \bottomrule
      \end{tabular}
    %   }
     \caption{\label{tab:sota_something} Results on SS V1.  ``N/A'' indicates the numbers are not available. ``Frames'' indicates the frames of a given input modality. ``views'' indicates  spatial crops $\times$ temporal clips. $\dagger$ denotes our reimplementation.
     }
\end{table*}

\subsection{Main Results}
In the following, we present the main results on multiple video benchmarks. To indicate the model that takes dynamic appearance as input, we prefix the model name with DA, and the configuration of $S \times T$ for generating the dynamic appearance is provided as a subscript beside ``DA'' (\ie DA$_{S \times T}$).\\

\subsubsection{Results on Something-Something}
Table~\ref{tab:sota_something} summarizes a comprehensive comparison, including the inference protocols, corresponding computational costs (FLOPs) and the prediction accuracy.
Our method surpasses the listed methods by good margins.
Our DA TSM taking dynamic appearance as input has 1.4\% and 10.6\% higher top-1 accuracy than TSM R50 taking RGB frame as input and TSM R50 taking optical flow as input, respectively.
Moreover, our DA TSM achieves similar accuracy (50.1\% \vs 50.6\%) to the ensemble model of TSM which fuses the predictions of the RGB frame and optical flow modalities.
However, the computation of our DA TSM is nearly half of TSM$_{\textup{RGB+Flow}}$, when the computation of its optical flow estimation has not been included.
The ViT-B~\cite{dosovitskiy2021image} with some small modifications described in~Sec.~\ref{sec: Action Recognition Pipeline} provides 45.7\% top-1 accuracy, which is lower than most listed models.
We give the reason that SS V1 requires models with strong temporal relation reasoning abilities, but the modified ViT-B does not model the interactions between different frames in its attention blocks. Nevertheless, our DA still contributes some improvement to ViT-B (+0.7\%).
our DA X3D-XS with X3D-XS backbone produces a similar accuracy as NL I3D~\cite{zhou2018temporal}, but requires 210 times fewer FLOPs per spacetime `view'. Meanwhile, DA X3D-XS also produces higher top-1 accuracy (+2.5\%) than X3D-XS. By adopting X3D-M as the backbone, our DA X3D-M achieves 53.7\% top-1 accuracy, which is 1.7\% higher than X3D-M and  1.4\% higher than TEA~\cite{li2020tea}.

\subsubsection{Results on K400, UCF101 and HMDB51}
Table~\ref{tab:sota_kinetics} shows the results on K400, where we list the models with the spatial input size smaller than $256^2$.
DA X3D-M achieves 77.3\% top-1 accuracy, which is better than X3D-M by 1.3\% and better than I3D~\cite{carreira2017quo} by a margin of +6.2\%.
The video-adapted ViT-B~\cite{dosovitskiy2021image} has 76.5\% top-1 accuracy which is 1.5\% and 2.1\% lower than the ViT-based architectures TimeSformer~\cite{bertasius2021space} and ViT-B-VTN~\cite{neimark2021video}, respectively. After taking the dynamic appearance as input, we improve the accuracy of ViT-B to 78.2\% (1.7\%), which is on par with other Transformer-based models. It is worth mentioning that our improvement in the Transformer backbone is higher than that in the CNN backbones. Although the scene information on K400 is widely considered to be important when distinguishing some actions, the improvement caused by filtering out the static appearance suggests that static features that do not change from frame to frame may impede the enhanced feature learning abilities in standard networks.

The results on UCF101~\cite{soomro2012ucf101} and HMDB51~\cite{kuehne2011hmdb} are shown in Table~\ref{tab: ucf_hmdb}, where we report the mean class accuracy over the three official splits. The recognition accuracy is already saturated on these relatively small-scale datasets. In order to avoid over-fitting, we pretrain our model on Kinetics. We consider the inference protocol (3 crops$\times$2 clips) for accuracy. DA TSM outperforms most other methods without employing tricks such as multi-stream fusion.

\begin{table*}
\centering
\begin{minipage}[t]{0.65\textwidth}

      \scalebox{0.86}{
      \setlength{\tabcolsep}{1.3pt}
      \begin{tabular}{l c c c c c}
      \toprule
      \bfseries Method & \bfseries pretrain & \tabincell{c}{\bfseries Frames \\ \bfseries $\times$ views} &  \bfseries FLOPs & \tabincell{c}{\bfseries Top1 \\ \bfseries (\%)} & \tabincell{c}{\bfseries Top5 \\ \bfseries (\%)} \\
      \midrule
      
      SlowFast$_{4\times16}$~\cite{feichtenhofer2019slowfast} & \multirow{5}*{IN-1K} & 32$\times$30 & 1083G & 75.6 & 92.1\\
      ip-CSN~\cite{tran2019video} &   & 32$\times$30& 2490G & 76.8 & 92.5\\

      I3D$_{\mathrm{RGB}}$~\cite{carreira2017quo} &  & 64$\times$N/A & N/A & 71.1 & 89.3\\
      NL I3D~\cite{wang2018non} &   & 128$\times$30 & 10770G & 77.7 & 93.3\\
      MViT-B$_{16\times4}$~\cite{fan2021multiscale} &   & 16$\times$5 & 352G& 78.4& 93.5 \\
      \cline{2-2}
      TimeSformer~\cite{bertasius2021space} & \multirow{2}*{IN-21K} & 8$\times$3 & 590G &  78.0 &  93.7\\
      ViT-B-VTN~\cite{neimark2021video} &  & 250$\times$1 & 4218G & 78.6& 93.7 \\
    %   TimeSformer~\cite{bertasius2021space} & IN-1K & 8$\times$3 & 590G & 75.8 & - \\
      \midrule
      \midrule
      TSM R50~\cite{lin2019tsm} & \multirow{2}*{IN-1K} & 16$\times$30 & 2577G & 74.7 & - \\
      \textbf{DA$_{16\times8}$ TSM} &  & 16$\times$30 & 2602G & 75.9 & 92.2 \\
      \midrule
    X3D-M~\cite{feichtenhofer2020x3d} & \multirow{2}*{None} & 16$\times$30 & 191G & 76.0 & 92.3\\
      \textbf{DA$_{16\times8}$X3D-M} &  & 16$\times$30 & 216G & 77.3 & 92.8 \\
      \midrule
      ViT-B$\dagger$~\cite{dosovitskiy2021image}  & \multirow{2}*{IN-21K} & 8$\times$3 & 808G & 76.5 & 92.6 \\
      \tabincell{l}{\bfseries DA$_{8\times8}$ViT-B}$\dagger$  & & 8$\times$3 & 809G & 78.2 & 93.4\\

    \bottomrule
      \end{tabular}
      }
  \makeatletter\def\@captype{table}\makeatother\caption{\label{tab:sota_kinetics} Results on K400. 
  We report the inference cost of multiple ``views'' (spatial crops × temporal clips).
  $\dagger$ denotes our reimplementation.}
\end{minipage}
\begin{minipage}[t]{0.001\textwidth}
\end{minipage}
\begin{minipage}[t]{0.32\textwidth}
\centering
  \scalebox{0.86}{
  \setlength{\tabcolsep}{1.3pt}
  \begin{tabular}{l c c}
  \\
  \toprule
  \bfseries Method  & \bfseries HMDB &  \bfseries UCF  \\ \midrule 
  StNet~\cite{he2019stnet}  & - & 93.5 \\
  TSM~\cite{lin2019tsm}  & 73.5 & 95.9 \\
  STM~\cite{jiang2019stm} & 72.2 & 96.2 \\ 
  TEA~\cite{li2020tea} & 73.3 & 96.9 \\
  Four-Stream~\cite{bilen2017action}  & 72.5 & 95.5 \\
  TVNet~\cite{fan2018end}   & 71.0 & 94.5 \\
  TSN$_{\textup{\tiny RGB+Flow}}$~\cite{wang2016temporal} & 68.5 & 94.0 \\
  OFF$_{\textup{\tiny RGB+Flow}}$~\cite{sun2018optical}  & 74.2 & 96.0 \\
  \midrule
 \textbf{DA$_{8\times16}$TSM}  & 75.3 & 96.5\\
 \bottomrule
  \\
 \\
 \\
 \\
 \\
  \end{tabular}}
  \makeatletter\def\@captype{table}\makeatother\caption{\label{tab: ucf_hmdb}Results on HMDB51 and UCF101. We report the mean class accuracy (\%) over the three official splits.}
\end{minipage}
\end{table*}

\subsection{Ablation Studies}
To investigate more configurations we employ the smaller X3D-XS~\cite{feichtenhofer2020x3d} as the backbone network processing $S=4$ segments, unless specified otherwise.

\subsubsection{Smaller Projection Error, Higher Classification Accuracy?}
\begin{figure}[!t]
\begin{center}
\resizebox{.48\textwidth}{!}{
\includegraphics[width=.6\textwidth]{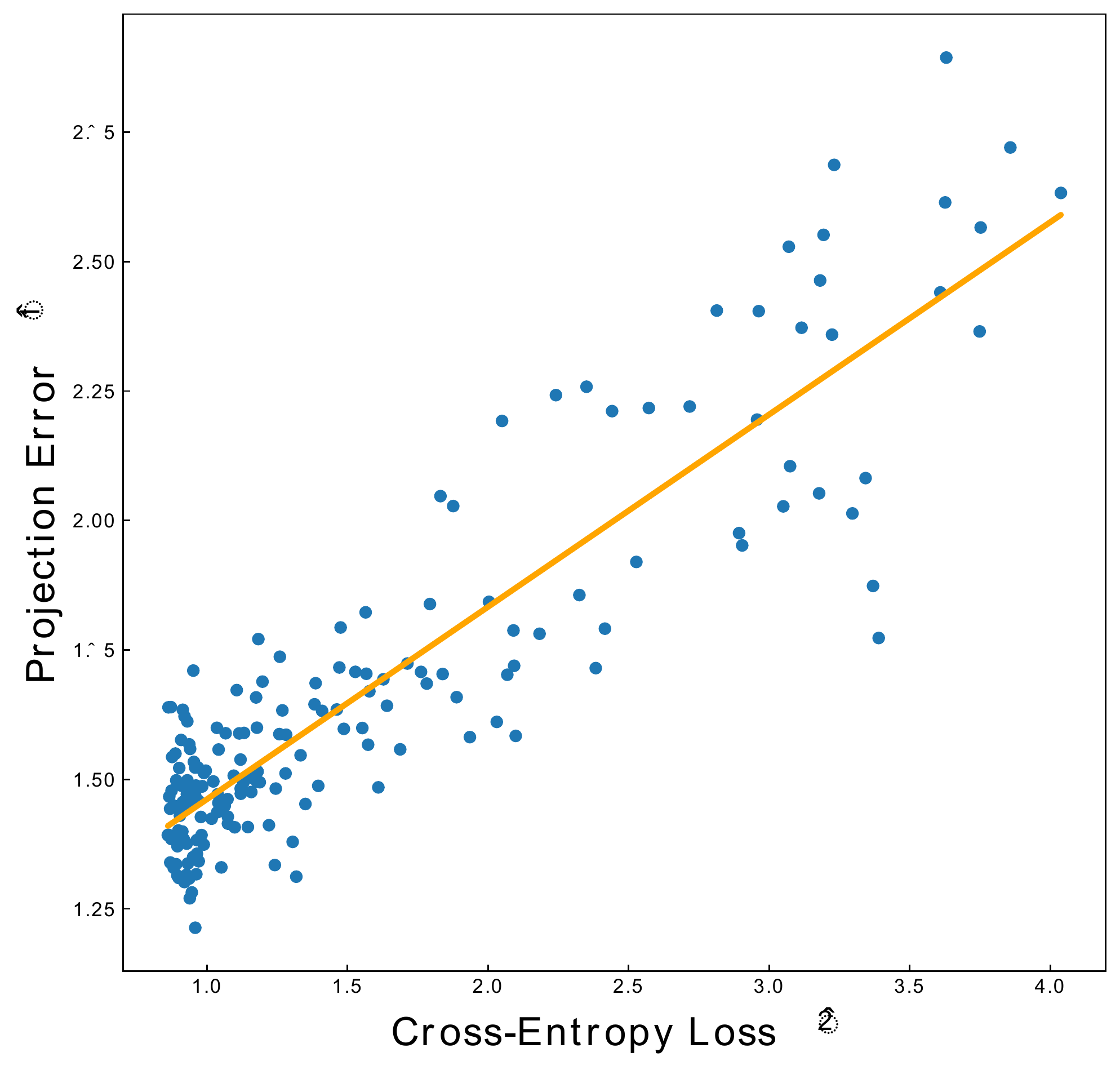} \hspace{1cm}
\includegraphics[width=.6\textwidth]{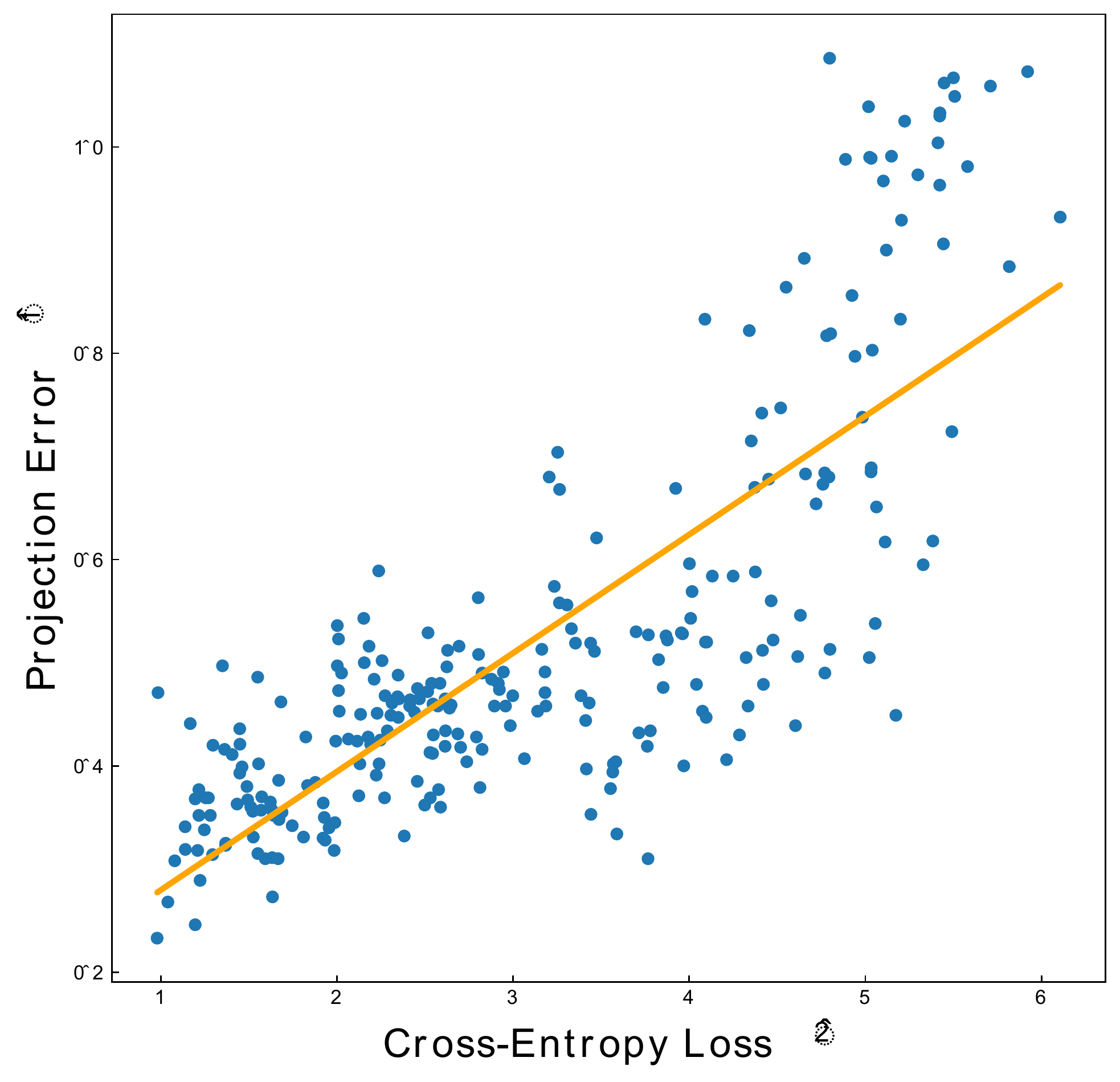}
}
\end{center}
\hspace{1.3cm} (a) UCF101  \hspace{2.2cm} (b) Mini-Kinetics
\caption{Example scatterplots of variables $\widehat{\mathcal{L}}^{1}$ and $\widehat{\mathcal{L}}^{2}$ on dataset (a) UCF101 and (b) Mini-Kinetics}
\label{fig: correlation_scatterplots}
\end{figure}

The criterion of the dynamic appearance learning task $\widehat{\mathcal{L}}^{1}(\Theta^1)$ and the action recognition criterion $\widehat{\mathcal{L}}^{2}(\Theta^1,\Theta^2)$ are supposed to be positive within a certain range, but this is not always true.
To find out the correlation of $\widehat{\mathcal{L}}^{1}$ and $\widehat{\mathcal{L}}^{2}$ we optimize the model with only $\widehat{\mathcal{L}}^{2}$ and record the changes for $\widehat{\mathcal{L}}^{1}$ and $\widehat{\mathcal{L}}^{2}$ in different training periods. From the plots in Fig.~\ref{fig: correlation_scatterplots}(a) and (b) on UCF101 and Mini-Kinetics, respectively, we observe that  $\widehat{\mathcal{L}}^{1}$ and $\widehat{\mathcal{L}}^{2}$ show positive correlation within the chosen range. The positive correlation indicates that it is feasible to set the projection task $\widehat{\mathcal{L}}^{1}(\Theta^1)$ of PWTP as an auxiliary task helping the model to generalize better on the action recognition task $\widehat{\mathcal{L}}^{2}(\Theta^1,\Theta^2)$. Outside of the specific range, we may obtain negative correlation, given the evidence from~Table~\ref{tab: joint_train_eva}, where separate training causes $\widehat{\mathcal{L}}^{1}$ to over-focus on the auxiliary task, resulting in a performance drop on the primary recognition task. When optimizing the PWTP module independently, a too small $\widehat{\mathcal{L}}^{1}$ could result in PWTP summarizing both static and dynamic appearance information into the same subspace. Meanwhile, the projection residual $\mathbf{p}$ would carry very little information, failing to represent abundant dynamic appearances. To tackle this issue, we set $D << T$, as stated in Sec.~\ref{sec:Pixel-Wise Temporal Projection}.

\begin{figure*}[!t]
\begin{center}
\resizebox{.8\textwidth}{!}{
\includegraphics{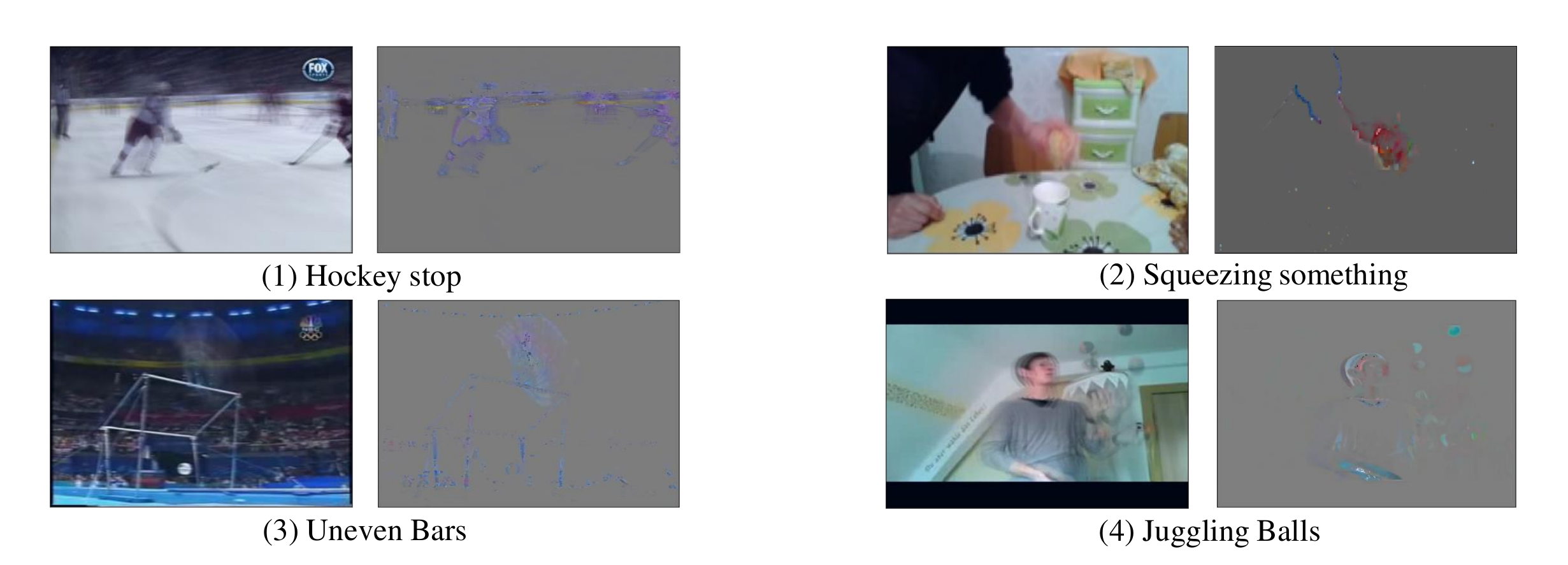} }
\end{center}
\caption{Example video sequences visualized as their frame averaging (left) and corresponding dynamic appearances (right). The videos are randomly picked from the four datasets.}
\label{fig: examples}
\end{figure*}

\begin{table*}[t!]
\centering
\captionsetup[subfloat]{labelformat=parens, labelsep=space, skip=7pt, position=bottom}

    \subfloat[Evaluating various settings of convolution $\mathcal{F}^{k \times k}_{s}$ in the PWTP module \label{tab: abla_PWTP_sconv}]{
    \scalebox{0.86}{
    \centering
    \setlength{\tabcolsep}{5pt}
    \begin{tabularx}{0.65\textwidth}{c c c c c c c c}
    \toprule
    Stride  &  Kernel size  & \#Channels  & \multicolumn{2}{c}{Mini-K} & \multicolumn{2}{c}{SS V1} & \multirow{2}*{FLOPs} \\
    \cmidrule(lr){4-5} \cmidrule(lr){6-7}
 $s$ &  $k \times k$ & $C'$ &  ENoPR  & Acc.   &   ENoPR & Acc.  &  \\
    \midrule
     1 &  $7\times 7$ & 12 & 0.26 & 55.1 & 0.14  & 40.9 & 2.60G \\
     1 &  $7\times 7$ & 24 & 0.22 & 58.6 & 0.13 & 41.1 & 4.06G \\
     4 &  $7\times 7$  & 24 & 0.32 & 58.1 & 0.18 & 41.3 & 0.95G \\
    \rowcolor{MyGray} 8 &  $9\times 9$  & 24 & 0.41 & 56.9&  0.27 & 41.0 & 0.82G \\
     8 &  $11\times 11$  & 24 & 0.37 & 57.1 & 0.20 & 41.2 & 0.86G \\
    \bottomrule
    \end{tabularx}}}
    \subfloat[Evaluating PWTP when changing $D$ and $T$ \label{tab: abla_PWTP_td}]{
    \scalebox{0.86}{
    \centering
    \setlength{\tabcolsep}{6pt}
    \begin{tabularx}{0.45\textwidth}{c c c c c c c}
    \toprule
    \multirow{2}*{$T$} &  \multirow{2}*{$D$}  & \multicolumn{2}{c}{Mini-K} & \multicolumn{2}{c}{SS V1} & \multirow{2}*{FLOPs} \\
    \cmidrule(lr){3-4} \cmidrule(lr){5-6}
 &  &  ENoPR  & Acc.   &   ENoPR  & Acc.  &  \\
    \midrule
     4 &  1 & 0.26  & 56.2 & 0.16 & 39.3 & 0.71G \\
     4 &  2 & 0.24 & 55.3 & 0.08 & 39.5 & 0.71G \\
     8 &  1 & 0.41  & 56.9 & 0.27 & 41.0 & 0.82G \\
     8 &  3 & 0.33  & 57.0 & 0.13 & 41.3 & 0.83G \\
     16 & 1 & 0.72  & \cellcolor{MyGray}58.6 & 0.39 & \cellcolor{MyGray} 42.1 & 1.12G \\
    \bottomrule
    \end{tabularx}}}
\caption{ Ablation Studies for PWTP on Something-Something (SS) V1 and Mini-Kinetics (Mini-K). The computational cost (FLOPs) of X3D-XS with a PWTP module embedded is reported.}
\end{table*}

\subsubsection{Joint training}
The dynamic appearance learning task and action recognition task could either be optimized separately or through joint training.  In Table~\ref{tab: joint_train_eva} we investigate the effect of separate training and different joint training approaches on performance. 
In summary, joint training approaches have higher accuracy than the separate training strategy. 
We note that the joint training with MGDA~\cite{desideri2012multiple} produces higher accuracy than the fixed scale strategy, which demonstrate the effectiveness of MGDA. We have also explored some other joint training approaches, the details of which are provided in the Appendix~D from the SM.

\begin{table}[!t]
    \centering
    \scalebox{.9}{
    \setlength{\tabcolsep}{19pt}
    \small
    \begin{tabular}{l c c c}
    \\
    \toprule
     \multirow{2}*{Joint Training Approach} & ENoPR  & Acc.  \\
     & ($\widehat{\mathcal{L}}^{1}$)& (\%) \\
    \midrule
    Baseline & - & 38.3 \\
    Separate training & 0.17 & 39.4 \\
    Constant Scale ($\alpha=0$)& 0.43 & 39.2   \\
    Constant Scale ($\alpha=0.5$)& 0.34 & 40.1  \\
    Constant Scale ($\alpha=0.9$) & 0.22 & 39.3  \\
    \rowcolor{MyGray}
    MGDA & 0.27 & 41.0  \\
    \bottomrule
    \end{tabular}
    }
        \caption{Results for different joint training approaches on SS V1. The baseline is the network with RGB frame input.}
    \label{tab: joint_train_eva}
    % %\vspace*{-0.2cm}
\end{table}

\begin{table}[t]
\small
% %\vspace*{-0.4cm}
\centering
\scalebox{.9}{
\setlength{\tabcolsep}{0pt}
\begin{tabular}{l c c c c}
\toprule
\multirow{2}*{ Representation Method} & \multicolumn{2}{c}{ Efficiency Metrics}  & \multirow{2}*{  SS V1} & \multirow{2}*{  UCF101}\\
\cmidrule{2-3}
 &  FLOPs &  \#Param. & & \\
\midrule
RGB (baseline) &- &-  & 46.5 & 87.1\\
Flow &- & -  & 37.4 & 88.5 \\
RGB+Flow & - & - & 49.8 & 93.9 \\
\midrule
RGB Diff~\cite{wang2016temporal}  &- & -  & 46.6 &  87.0 \\
Dynamic Image$\dagger$~\cite{bilen2017action} & - & -  & 43.3 & 86.2 \\
FlowNetC$\dagger$~\cite{ilg2017flownet}  & 444G & 39.2M   & 26.3 & 87.3 \\
FlowNetS$\dagger$~\cite{ilg2017flownet}  & 356G &  38.7M   & 23.4 & 86.8 \\
TVNet$\dagger$~\cite{fan2018end} & 3.30G & 0.20K & 45.2  & 88.6  \\
\midrule
\textbf{DA} & 0.23G & 7.13K  & 48.7  & 89.7 \\
% \textbf{RGB+DA} & - & - & {\bf 52.9}  & \textbf{90.2}\\
\bottomrule
\end{tabular}
}
\caption{DA \vs other motion representation methods.  $\dagger$ denotes our reimplementation. The additional parameters and computation (FLOPs) required by the representation methods are reported.}
\label{table:compar_reps}
\end{table}

\subsubsection{Instantiations and Settings of PWTP}
\label{abl_PWTP}
We consider various configurations of PWTP on SS V1 and Mini-Kinetics, vary $T$ and $D$ from Eq.~\eqref{eq:mlp_resahpe} and evaluate the recognition accuracy. Table~\ref{tab: abla_PWTP_td} shows that the configurations with $T=8$ have higher accuracy than $T=4$ but lower accuracy than $T=16$, which suggests that longer frame sequences can help PWTP to capture a diversity of dynamic appearances regarding movement and generate better results. When fixing $T=8$, the setting of $D=3$ is better than that of $D=1$, which suggests that a relatively larger $D$ endues PWTP with stronger information summarization capacity and helps generate lower ENoPR and higher accuracy. Nevertheless, $D$ should not set to be larger than half of $T$, as we have seen a performance drop when setting $D>\frac{1}{2}T$ in our preliminary experiments.

The spatial convolution $\mathcal{F}^{k \times k}_{s}(\cdot; \theta)$ from Eq.~\eqref{eq_conv1} is used for local information aggregation. We ablate various settings of $\mathcal{F}^{k \times k}_{s}(\cdot; \theta)$ , including changing the stride $s$, kernel size $k \times k$, output channel number $C'$ and the results are provided in~Table~\ref{tab: abla_PWTP_sconv}. 
A larger kernel of $\mathcal{F}^{k \times k}_{s}(\cdot; \theta)$ can receive information from a larger local region, and therefore PWTP can encode additional spatial information into the temporal descriptor $\mathbf{u}$, which helps the model to produce higher accuracy. In addition, Table~\ref{tab: abla_PWTP_sconv} shows that a large stride $s$ for the convolution $\mathcal{F}^{k \times k}_{s}(\cdot; \theta)$ can significantly reduce the computational cost while maintaining performance.

\subsubsection{Efficiency and Effectiveness of the Dynamic Appearance}
We draw an apple-to-apple comparison between the proposed dynamic appearance and other motion representations \cite{bilen2017action,fan2018end,ilg2017flownet,zach2007duality}. The comparison results are shown in Table~\ref{table:compar_reps}. For all the video representation methods, we guarantee they consume the same number of raw video frames (48 frames). More details can be found in Appendix~B from the SM. The motion representations produced by these methods are used as inputs to TSM R50~\cite{lin2019tsm}. The prediction scores are obtained by the average consensus of eight temporal segments~\cite{wang2016temporal}. The network that takes the proposed dynamic appearance as input outperforms all other motion representation methods by big margins within reasonable computational costs. 
In SS V1, our method achieves similar accuracy to the fusion of ``RGB+Flow''.

\subsection{Visualization analysis}
\begin{figure}[!t]
    % \centering
    \resizebox{.48\textwidth}{!}{
    \includegraphics{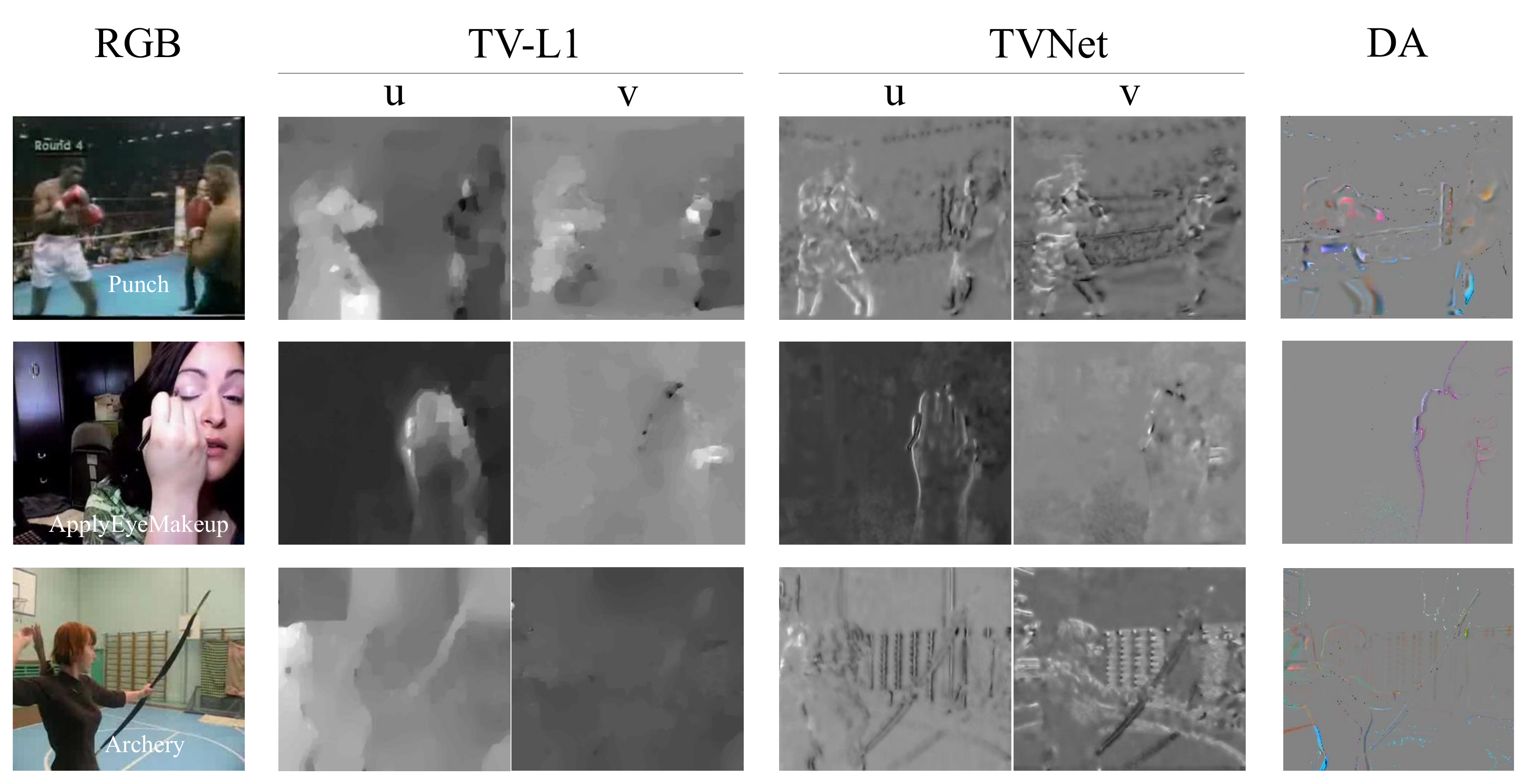}}
    \caption{Visual examples of different video representations. Best viewed in color and zoomed in.}
    \label{fig:cmpar_other_reps}
\end{figure}

\begin{figure}[!t]
\begin{center}
\resizebox{.47\textwidth}{!}{
\includegraphics{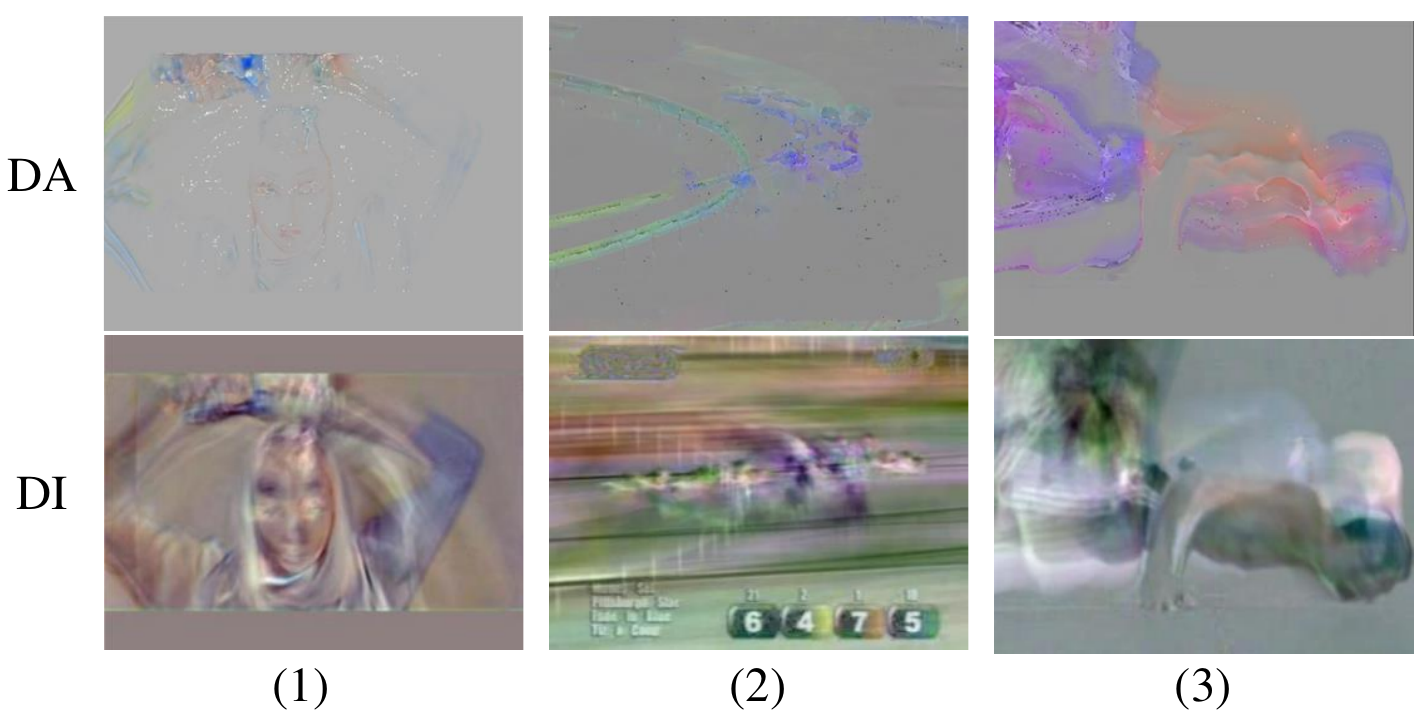} }
\end{center}
\caption{Visualization of Dynamic Appearance (DA) and Dynamic Image (DI).}
\label{fig: compar_da_di}
\end{figure}
Visualization results of dynamic appearances are provided in Fig.~\ref{fig: examples}. the example dynamic appearance suppresses the background movement and stationary information and retains the appearance information characteristic to motion.
In Fig.~\ref{fig:cmpar_other_reps} we compare dynamic appearances with TV-L1 Flow~\cite{zach2007duality} and TVNet~\cite{fan2018end}. TV-L1 Flow and TVNet represent motion with instantaneous image velocities, losing the color and appearance information related to moving objects. However, the dynamic appearances preserved the essential parts of moving objects' visual information, which is vital for discriminating different actions. 
In Fig.~\ref{fig: compar_da_di},  we compare visually the Dynamic Appearance and Dynamic Image~\cite{bilen2017action}. 
The dynamic image results in poor discrimination between the moving objects and background when we have camera shaking or movement, as we can observe from Fig.~\ref{fig: compar_da_di}(2). On the contrary, the dynamic appearance is robust to camera movement, clearly representing the visual information related to motion. More visualization examples are provided in Appendix~E, where we show that Dynamic Appearance steadily works on a wide range of actions.

\section{Conclusion}

We have proposed the Pixel-Wise Temporal Projection (PWTP), a lightweight and backbone-agnostic module, which achieves the disentanglement of video static and dynamic appearance. The extracted dynamic appearance (DA) characterizes movement by summarizing the visual appearance information that changes from frame to frame. We integrate the PWTP module with a CNN or Visual Transformer (ViT) into an efficient and effective spatio-temporal architecture, set in an end-to-end trainable framework. The multiple gradient descent algorithm is used for the joint training of PWTP and the deep network. With extensive experimental results on multiple challenging video benchmarks, we have demonstrated that the proposed dynamic appearances are a qualified input resource to deep networks, showing great advantages over RGB frames and the optical flow inputs, in terms of efficiency and effectiveness.
The proposed video representation methodology can contribute to designing optimal spatio-temporal video modeling systems.

% Use \bibliography{yourbibfile} instead or the References section will not appear in your paper
\bibliography{main.bbl}

\begin{thebibliography}{57}
\providecommand{\natexlab}[1]{#1}

\bibitem[{Ahad et~al.(2008)Ahad, Ogata, Tan, Kim, and
  Ishikawa}]{ahad2008motion}
Ahad, M.; Ogata, T.; Tan, J.; Kim, H.; and Ishikawa, S. 2008.
\newblock Motion recognition approach to solve overwriting in complex actions.
\newblock In \emph{International Conference on Automatic Face \& Gesture
  Recognition}, 1--6. IEEE.

\bibitem[{Arnab et~al.(2021)Arnab, Dehghani, Heigold, Sun, Lu{\v{c}}i{\'c}, and
  Schmid}]{arnab2021vivit}
Arnab, A.; Dehghani, M.; Heigold, G.; Sun, C.; Lu{\v{c}}i{\'c}, M.; and Schmid,
  C. 2021.
\newblock Vivit: A video vision transformer.
\newblock In \emph{Proc. IEEE Int. Conference Computer Vision (ICCV)},
  6836--6846.

\bibitem[{Bertasius, Wang, and Torresani(2021)}]{bertasius2021space}
Bertasius, G.; Wang, H.; and Torresani, L. 2021.
\newblock Is space-time attention all you need for video understanding.
\newblock In \emph{Proc. Int. Conference Mach. Learn. (ICML)}.

\bibitem[{Bilen et~al.(2018)Bilen, Fernando, Gavves, and
  Vedaldi}]{bilen2017action}
Bilen, H.; Fernando, B.; Gavves, E.; and Vedaldi, A. 2018.
\newblock Action recognition with dynamic image networks.
\newblock \emph{IEEE Trans. on Pattern Analysis and Machine Intelligence},
  40(12): 2799--2813.

\bibitem[{Bradski and Davis(2002)}]{bradski2002motion}
Bradski, G.; and Davis, J. 2002.
\newblock Motion segmentation and pose recognition with motion history
  gradients.
\newblock \emph{Machine Vision and Applications}, 13(3): 174--184.

\bibitem[{Carreira and Zisserman(2017)}]{carreira2017quo}
Carreira, J.; and Zisserman, A. 2017.
\newblock Quo vadis, action recognition? a new model and the kinetics dataset.
\newblock In \emph{Proc. IEEE Conference Computer Vision Pattern Recog.
  (CVPR)}, 4724--4733.

\bibitem[{Deng et~al.(2009)Deng, Dong, Socher, Li, Li, and
  Fei-Fei}]{imagenet_cvpr09}
Deng, J.; Dong, W.; Socher, R.; Li, L.-J.; Li, K.; and Fei-Fei, L. 2009.
\newblock Imagenet: A large-scale hierarchical image database.
\newblock In \emph{Proc. IEEE Conference Computer Vision Pattern Recog.
  (CVPR)}, 248--255.

\bibitem[{D{\'e}sid{\'e}ri(2012)}]{desideri2012multiple}
D{\'e}sid{\'e}ri, J. 2012.
\newblock Multiple-gradient descent algorithm ({MGDA}) for multiobjective
  optimization.
\newblock \emph{Comptes Rendus Mathematique}, 350(5-6): 313--318.

\bibitem[{Dosovitskiy et~al.(2021)Dosovitskiy, Beyer, Kolesnikov, Weissenborn,
  Zhai, Unterthiner, Dehghani, Minderer, Heigold, Gelly
  et~al.}]{dosovitskiy2021image}
Dosovitskiy, A.; Beyer, L.; Kolesnikov, A.; Weissenborn, D.; Zhai, X.;
  Unterthiner, T.; Dehghani, M.; Minderer, M.; Heigold, G.; Gelly, S.; et~al.
  2021.
\newblock An image is worth 16x16 words: Transformers for image recognition at
  scale.
\newblock \emph{Int. Conf. Learn. Represent. (ICLR), arXiv preprint
  arXiv:2010.11929}.

\bibitem[{Dosovitskiy et~al.(2015)Dosovitskiy, Fischer, Ilg, Hausser, Hazirbas,
  Golkov, Van Der~Smagt, Cremers, and Brox}]{dosovitskiy2015flownet}
Dosovitskiy, A.; Fischer, P.; Ilg, E.; Hausser, P.; Hazirbas, C.; Golkov, V.;
  Van Der~Smagt, P.; Cremers, D.; and Brox, T. 2015.
\newblock Flownet: Learning optical flow with convolutional networks.
\newblock In \emph{Proc. IEEE Int. Conference Computer Vision (ICCV)},
  2758--2766.

\bibitem[{Fan et~al.(2021)Fan, Xiong, Mangalam, Li, Yan, Malik, and
  Feichtenhofer}]{fan2021multiscale}
Fan, H.; Xiong, B.; Mangalam, K.; Li, Y.; Yan, Z.; Malik, J.; and
  Feichtenhofer, C. 2021.
\newblock Multiscale vision transformers.
\newblock In \emph{Proc. IEEE Int. Conference Computer Vision (ICCV)},
  6824--6835.

\bibitem[{Fan et~al.(2018)Fan, Huang, Gan, Ermon, Gong, and Huang}]{fan2018end}
Fan, L.; Huang, W.; Gan, C.; Ermon, S.; Gong, B.; and Huang, J. 2018.
\newblock End-to-end learning of motion representation for video understanding.
\newblock In \emph{Proc. IEEE Conference Computer Vision Pattern Recog.
  (CVPR)}, 6016--6025.

\bibitem[{Feichtenhofer(2020)}]{feichtenhofer2020x3d}
Feichtenhofer, C. 2020.
\newblock {X3D}: Expanding Architectures for Efficient Video Recognition.
\newblock In \emph{Proc. IEEE Conf. Computer Vision and Pattern Recognition
  (CVPR)}, 203--213.

\bibitem[{Feichtenhofer et~al.(2019)Feichtenhofer, Fan, Malik, and
  He}]{feichtenhofer2019slowfast}
Feichtenhofer, C.; Fan, H.; Malik, J.; and He, K. 2019.
\newblock Slowfast networks for video recognition.
\newblock In \emph{Proc. IEEE Int. Conf. on Computer Vision (ICCV)},
  6202--6211.

\bibitem[{Feichtenhofer, Pinz, and
  Wildes(2016)}]{feichtenhofer2016spatiotemporal}
Feichtenhofer, C.; Pinz, A.; and Wildes, R. 2016.
\newblock Spatiotemporal residual networks for video action recognition.
\newblock In \emph{Advances in Neural Information Processing Systems (NIPS)},
  3468--3476.

\bibitem[{Feichtenhofer, Pinz, and
  Zisserman(2016)}]{feichtenhofer2016convolutional}
Feichtenhofer, C.; Pinz, A.; and Zisserman, A. 2016.
\newblock Convolutional two-stream network fusion for video action recognition.
\newblock In \emph{Proc. IEEE Conference Computer Vision Pattern Recog.
  (CVPR)}, 1933--1941.

\bibitem[{Fernando et~al.(2015)Fernando, Gavves, Oramas, Ghodrati, and
  Tuytelaars}]{fernando2015modeling}
Fernando, B.; Gavves, E.; Oramas, J.~M.; Ghodrati, A.; and Tuytelaars, T. 2015.
\newblock Modeling video evolution for action recognition.
\newblock In \emph{Proc. IEEE Conference Computer Vision Pattern Recog.
  (CVPR)}, 5378--5387.

\bibitem[{Goyal et~al.(2017)Goyal, Kahou, Michalski, Materzynska, Westphal,
  Kim, Haenel, Fruend, Yianilos, Mueller-Freitag et~al.}]{goyal2017something}
Goyal, R.; Kahou, S.~E.; Michalski, V.; Materzynska, J.; Westphal, S.; Kim, H.;
  Haenel, V.; Fruend, I.; Yianilos, P.; Mueller-Freitag, M.; et~al. 2017.
\newblock The" Something Something" Video Database for Learning and Evaluating
  Visual Common Sense.
\newblock In \emph{Proc. IEEE Int. Conf. on Computer Vision (ICCV)}, volume~1,
  5842--5850.

\bibitem[{Hara, Kataoka, and Satoh(2018)}]{hara2018can}
Hara, K.; Kataoka, H.; and Satoh, Y. 2018.
\newblock Can spatiotemporal 3d cnns retrace the history of 2d cnns and
  imagenet?
\newblock In \emph{Proc. IEEE Conference Computer Vision Pattern Recog.
  (CVPR)}, 6546--6555.

\bibitem[{He et~al.(2019)He, Zhou, Gan, Li, Liu, Li, Wang, and
  Wen}]{he2019stnet}
He, D.; Zhou, Z.; Gan, C.; Li, F.; Liu, X.; Li, Y.; Wang, L.; and Wen, S. 2019.
\newblock Stnet: Local and global spatial-temporal modeling for action
  recognition.
\newblock In \emph{Proc. AAAI Conference on Artif. Intel.}, volume~33,
  8401--8408.

\bibitem[{He et~al.(2016)He, Zhang, Ren, and Sun}]{he2016deep}
He, K.; Zhang, X.; Ren, S.; and Sun, J. 2016.
\newblock Deep residual learning for image recognition.
\newblock In \emph{Proc. IEEE Conference Computer Vision Pattern Recog.
  (CVPR)}, 770--778.

\bibitem[{Ilg et~al.(2017)Ilg, Mayer, Saikia, Keuper, Dosovitskiy, and
  Brox}]{ilg2017flownet}
Ilg, E.; Mayer, N.; Saikia, T.; Keuper, M.; Dosovitskiy, A.; and Brox, T. 2017.
\newblock Flownet 2.0: Evolution of optical flow estimation with deep networks.
\newblock In \emph{Proc. IEEE Conference Computer Vision Pattern Recog.
  (CVPR)}, volume~2, 2462--2470.

\bibitem[{Ioffe and Szegedy(2015)}]{szegedy2015bn}
Ioffe, S.; and Szegedy, C. 2015.
\newblock Batch Normalization: Accelerating Deep Network Training by Reducing
  Internal Covariate Shift.
\newblock In \emph{Proc. Int. Conference Mach. Learn. (ICML), vol. PMLR 37},
  448–456.

\bibitem[{Jiang et~al.(2019)Jiang, Wang, Gan, Wu, and Yan}]{jiang2019stm}
Jiang, B.; Wang, M.; Gan, W.; Wu, W.; and Yan, J. 2019.
\newblock Stm: Spatiotemporal and motion encoding for action recognition.
\newblock In \emph{Proc. IEEE Int. Conference Computer Vision (ICCV)},
  2000--2009.

\bibitem[{Karpathy et~al.(2014)Karpathy, Toderici, Shetty, Leung, Sukthankar,
  and Fei-Fei}]{karpathy2014large}
Karpathy, A.; Toderici, G.; Shetty, S.; Leung, T.; Sukthankar, R.; and Fei-Fei,
  L. 2014.
\newblock Large-scale video classification with convolutional neural networks.
\newblock In \emph{Proc. IEEE Conference Computer Vision Pattern Recog.
  (CVPR)}, 1725--1732.

\bibitem[{Krizhevsky, Sutskever, and Hinton(2012)}]{krizhevsky2012imagenet}
Krizhevsky, A.; Sutskever, I.; and Hinton, G. 2012.
\newblock Imagenet classification with deep convolutional neural networks.
\newblock In \emph{Advances Neural Information Process. Systems (NIPS)},
  1097--1105.

\bibitem[{Kuehne et~al.(2011)Kuehne, Jhuang, Garrote, Poggio, and
  Serre}]{kuehne2011hmdb}
Kuehne, H.; Jhuang, H.; Garrote, E.; Poggio, T.; and Serre, T. 2011.
\newblock {HMDB}: a large video database for human motion recognition.
\newblock In \emph{Proc. IEEE Int. Conference Computer Vision (ICCV)},
  2556--2563.

\bibitem[{Li et~al.(2020)Li, Ji, Shi, Zhang, Kang, and Wang}]{li2020tea}
Li, Y.; Ji, B.; Shi, X.; Zhang, J.; Kang, B.; and Wang, L. 2020.
\newblock TEA: Temporal Excitation and Aggregation for Action Recognition.
\newblock In \emph{Proc. IEEE Conference Computer Vision Pattern Recog.
  (CVPR)}, 909--918.

\bibitem[{Lin, Gan, and Han(2019)}]{lin2019tsm}
Lin, J.; Gan, C.; and Han, S. 2019.
\newblock Tsm: Temporal shift module for efficient video understanding.
\newblock In \emph{Proc. IEEE Int. Conference Computer Vision (ICCV)},
  7083--7093.

\bibitem[{Loshchilov and Hutter(2017)}]{loshchilov2016sgdr}
Loshchilov, I.; and Hutter, F. 2017.
\newblock {SGDR}: Stochastic gradient descent with warm restarts.
\newblock \emph{Int. Conf. Learn. Represent. (ICLR), arXiv preprint
  arXiv:1608.03983}.

\bibitem[{Loshchilov and Hutter(2019)}]{loshchilov2018decoupled}
Loshchilov, I.; and Hutter, F. 2019.
\newblock Decoupled Weight Decay Regularization.
\newblock In \emph{Int. Conf. Learn. Represent. (ICLR), arXiv preprint
  arXiv:1711.05101}.

\bibitem[{Neimark et~al.(2021)Neimark, Bar, Zohar, and
  Asselmann}]{neimark2021video}
Neimark, D.; Bar, O.; Zohar, M.; and Asselmann, D. 2021.
\newblock Video transformer network.
\newblock In \emph{Proc. IEEE Int. Conference Computer Vision (ICCV)},
  3163--3172.

\bibitem[{Ng et~al.(2018)Ng, Choi, Neumann, and Davis}]{ng2018actionflownet}
Ng, J. Y.-H.; Choi, J.; Neumann, J.; and Davis, L.~S. 2018.
\newblock Actionflownet: Learning motion representation for action recognition.
\newblock In \emph{2018 IEEE Winter Conference on Applications of Computer
  Vision (WACV)}, 1616--1624. IEEE.

\bibitem[{Piergiovanni and Ryoo(2019)}]{piergiovanni2019representation}
Piergiovanni, A.; and Ryoo, S. 2019.
\newblock Representation flow for action recognition.
\newblock In \emph{Proc. IEEE Conference Computer Vision Pattern Recog.
  (CVPR)}, 9945--9953.

\bibitem[{Qiu, Yao, and Mei(2017)}]{qiu2017learning}
Qiu, Z.; Yao, T.; and Mei, T. 2017.
\newblock Learning spatio-temporal representation with pseudo-3d residual
  networks.
\newblock In \emph{Proc. IEEE Int. Conference Computer Vision (ICCV)},
  5533--5541.

\bibitem[{Sener and Koltun(2018)}]{sener2018multi}
Sener, O.; and Koltun, V. 2018.
\newblock Multi-Task Learning as Multi-Objective Optimization.
\newblock In \emph{Advances Neural Information Process. Systems (NIPS)},
  525--536.

\bibitem[{Simonyan and Zisserman(2014)}]{simonyan2014two}
Simonyan, K.; and Zisserman, A. 2014.
\newblock Two-stream convolutional networks for action recognition in videos.
\newblock In \emph{Advances Neural Information Process. Systems (NIPS)},
  568--576.

\bibitem[{Simonyan and Zisserman(2015)}]{simonyan2014very}
Simonyan, K.; and Zisserman, A. 2015.
\newblock Very Deep Convolutional Networks for Large-Scale Image Recognition.
\newblock In \emph{Int. Conf. Learn. Represent. (ICLR), arXiv preprint
  arXiv:1409.1556}.

\bibitem[{Soomro, Zamir, and Shah(2012)}]{soomro2012ucf101}
Soomro, K.; Zamir, A.~R.; and Shah, M. 2012.
\newblock {UCF101}: A dataset of 101 human actions classes from videos in the
  wild.
\newblock \emph{arXiv preprint arXiv:1212.0402}.

\bibitem[{Sun et~al.(2021)Sun, Vlasic, Herrmann, Jampani, Krainin, Chang,
  Zabih, Freeman, and Liu}]{sun2021autoflow}
Sun, D.; Vlasic, D.; Herrmann, C.; Jampani, V.; Krainin, M.; Chang, H.; Zabih,
  R.; Freeman, W.; and Liu, C. 2021.
\newblock Autoflow: Learning a better training set for optical flow.
\newblock In \emph{Proceedings of the IEEE/CVF Conference on Computer Vision
  and Pattern Recognition (CVPR)}, 10093--10102.

\bibitem[{Sun et~al.(2018)Sun, Kuang, Sheng, Ouyang, and
  Zhang}]{sun2018optical}
Sun, S.; Kuang, Z.; Sheng, L.; Ouyang, W.; and Zhang, W. 2018.
\newblock Optical flow guided feature: A fast and robust motion representation
  for video action recognition.
\newblock In \emph{Proc. IEEE Conference Computer Vision Pattern Recog.
  (CVPR)}, 1390--1399.

\bibitem[{Szegedy et~al.(2015)Szegedy, Liu, Jia, Sermanet, Reed, Anguelov,
  Erhan, Vanhoucke, and Rabinovich}]{szegedy2015going}
Szegedy, C.; Liu, W.; Jia, Y.; Sermanet, P.; Reed, S.; Anguelov, D.; Erhan, D.;
  Vanhoucke, V.; and Rabinovich, A. 2015.
\newblock Going deeper with convolutions.
\newblock In \emph{Proc. IEEE Conference Computer Vision Pattern Recog.
  (CVPR)}, 1--9.

\bibitem[{Szegedy et~al.(2016)Szegedy, Vanhoucke, Ioffe, Shlens, and
  Wojna}]{szegedy2016rethinking}
Szegedy, C.; Vanhoucke, V.; Ioffe, S.; Shlens, J.; and Wojna, Z. 2016.
\newblock Rethinking the inception architecture for Comp. vision.
\newblock In \emph{Proc. IEEE Conference Computer Vision Pattern Recog.
  (CVPR)}, 2818--2826.

\bibitem[{Taylor et~al.(2010)Taylor, Fergus, LeCun, and
  Bregler}]{taylor2010convolutional}
Taylor, G.~W.; Fergus, R.; LeCun, Y.; and Bregler, C. 2010.
\newblock Convolutional learning of spatio-temporal features.
\newblock In \emph{Proc. European Conference Computer Vision (ECCV)}, 140--153.

\bibitem[{Tran et~al.(2015)Tran, Bourdev, Fergus, Torresani, and
  Paluri}]{tran2015learning}
Tran, D.; Bourdev, L.; Fergus, R.; Torresani, L.; and Paluri, M. 2015.
\newblock Learning spatiotemporal features with {3D} convolutional networks.
\newblock In \emph{Proc. IEEE Int. Conference Computer Vision (ICCV)},
  4489--4497.

\bibitem[{Tran et~al.(2019)Tran, Wang, Torresani, and Feiszli}]{tran2019video}
Tran, D.; Wang, H.; Torresani, L.; and Feiszli, M. 2019.
\newblock Video classification with channel-separated convolutional networks.
\newblock In \emph{Proc. IEEE Int. Conference Computer Vision (ICCV)},
  5552--5561.

\bibitem[{Tran et~al.(2018)Tran, Wang, Torresani, Ray, LeCun, and
  Paluri}]{tran2018closer}
Tran, D.; Wang, H.; Torresani, L.; Ray, J.; LeCun, Y.; and Paluri, M. 2018.
\newblock A closer look at spatiotemporal convolutions for action recognition.
\newblock In \emph{Proc. IEEE Conference Computer Vision Pattern Recog.
  (CVPR)}, 6450--6459.

\bibitem[{Wang et~al.(2018{\natexlab{a}})Wang, Cherian, Porikli, and
  Gould}]{wang2018video}
Wang, J.; Cherian, A.; Porikli, F.; and Gould, S. 2018{\natexlab{a}}.
\newblock Video representation learning using discriminative pooling.
\newblock In \emph{Proc. IEEE Conference Computer Vision Pattern Recog.
  (CVPR)}, 1149--1158.

\bibitem[{Wang et~al.(2021)Wang, Tong, Ji, and Wu}]{wang2021tdn}
Wang, L.; Tong, Z.; Ji, B.; and Wu, G. 2021.
\newblock Tdn: Temporal difference networks for efficient action recognition.
\newblock In \emph{Proc. IEEE Conference Computer Vision Pattern Recog.
  (CVPR)}, 1895--1904.

\bibitem[{Wang et~al.(2016)Wang, Xiong, Wang, Qiao, Lin, Tang, and
  Van~Gool}]{wang2016temporal}
Wang, L.; Xiong, Y.; Wang, Z.; Qiao, Y.; Lin, D.; Tang, X.; and Van~Gool, L.
  2016.
\newblock Temporal segment networks: Towards good practices for deep action
  recognition.
\newblock In \emph{Proc. European Conference Computer Vision (ECCV), vol LNCS
  9912}, 20--36.

\bibitem[{Wang et~al.(2018{\natexlab{b}})Wang, Girshick, Gupta, and
  He}]{wang2018non}
Wang, X.; Girshick, R.; Gupta, A.; and He, K. 2018{\natexlab{b}}.
\newblock Non-local neural networks.
\newblock In \emph{Proc. IEEE Conference Computer Vision Pattern Recog.
  (CVPR)}, 7794--7803.

\bibitem[{Wang et~al.(2017)Wang, Long, Wang, and Yu}]{wang2017spatiotemporal}
Wang, Y.; Long, M.; Wang, J.; and Yu, P. 2017.
\newblock Spatiotemporal pyramid network for video action recognition.
\newblock In \emph{Proc. IEEE Conference Computer Vision Pattern Recog.
  (CVPR)}, 1529--1538.

\bibitem[{Xie et~al.(2018)Xie, Sun, Huang, Tu, and Murphy}]{xie2018rethinking}
Xie, S.; Sun, C.; Huang, J.; Tu, Z.; and Murphy, K. 2018.
\newblock Rethinking spatiotemporal feature learning: Speed-accuracy trade-offs
  in video classification.
\newblock In \emph{Proc. European Conference Computer Vision (ECCV), vol. LNCS
  11219}, 305--321.

\bibitem[{Yue-Hei~Ng et~al.(2015)Yue-Hei~Ng, Hausknecht, Vijayanarasimhan,
  Vinyals, Monga, and Toderici}]{yue2015beyond}
Yue-Hei~Ng, J.; Hausknecht, M.; Vijayanarasimhan, S.; Vinyals, O.; Monga, R.;
  and Toderici, G. 2015.
\newblock Beyond short snippets: Deep networks for video classification.
\newblock In \emph{Proc. IEEE Conference Computer Vision Pattern Recog.
  (CVPR)}, 4694--4702.

\bibitem[{Zach, Pock, and Bischof(2007)}]{zach2007duality}
Zach, C.; Pock, T.; and Bischof, H. 2007.
\newblock A duality based approach for realtime TV-L 1 optical flow.
\newblock In \emph{Proc. Joint Pattern Recog. Symp., vol. LNCS 4713}, 214--223.

\bibitem[{Zhou et~al.(2018)Zhou, Andonian, Oliva, and
  Torralba}]{zhou2018temporal}
Zhou, B.; Andonian, A.; Oliva, A.; and Torralba, A. 2018.
\newblock Temporal relational reasoning in videos.
\newblock In \emph{Proc. European Conference Computer Vision (ECCV)}, 803--818.

\bibitem[{Zhu et~al.(2018)Zhu, Lan, Newsam, and Hauptmann}]{zhu2018hidden}
Zhu, Y.; Lan, Z.; Newsam, S.; and Hauptmann, A. 2018.
\newblock Hidden two-stream convolutional networks for action recognition.
\newblock In \emph{Asian Conference on Computer Vision}, 363--378. Springer.

\end{thebibliography}

\newpage

\appendix

\section{Additional Implementation details}
In the following we provide additional implementation details and parameter settings in our models.
We train our models using synchronized SGD with momentum 0.9 and a cosine learning rate schedule.
By default, we add a dropout layer before the classification layer in our models to prevent overfitting. To avoid over-fitting, we apply L2 regularization to the convolutional and classification layers. Meanwhile, we use label-smoothing~\cite{szegedy2016rethinking} of 0.1 and gradient clip of 20. Following the experimental settings in~\cite{lin2019tsm,wang2016temporal}, the learning rate and weight decay parameters for the classification layers are 5 times larger than those for the convolutional layers. Besides, when training on Something-Something V1, we do not apply random horizontal flip to the training data.

For efficient inference, we sample a single clip per video with center cropping used in the ablation studies.
The spatial size of a center crop is of $160\times160$ for the models with X3D-XS backbone, while for the others is $224\times224$.
When pursuing high accuracy, we sample multiple clips\&crops from the video and average the softmax scores of multiple spacetime ``views'' (spatial crops$\times$temporal clips) for prediction. Following the practice in ~\cite{feichtenhofer2019slowfast}, we approximate the fully-convolutional testing by taking 3 crops of $256\times256$ pixels ($160\times160$ for X3D-XS) to cover the spatial dimensions.

\subsubsection{Hyperparameters for the model based on ResNet}
The backbone networks are pretrained on ImageNet-1K~\cite{imagenet_cvpr09}.
On Kinetics400, we train our models on 64 GPUs (NVIDIA Tesla V100).
The initial learning rate, batch size, total epochs, weight decay and dropout ratio are set to 0.08, 512 (8 samples per GPU), 100, 2e-4 and 0.5, respectively. 
On Something-Something V1, we train our models in 32 GPUs, and the hyperparameters mentioned above are set to 0.12, 256 (8 samples per GPU), 50, 8e-4 and 0.8, respectively. We use linear warm-up~\cite{loshchilov2016sgdr} for the first 7 epochs to overcome early optimization difficulty.
When fine-tuning the Kinetics models on UCF101 and HMDB51, we train our models on 16 GPUs and freeze the entire batch normalization~\cite{szegedy2015bn} layers except for the first one to avoid overfitting, following the recipe in~\cite{wang2016temporal}.
The initial learning rate, batch size, total epochs, weight decay and dropout ratio are set to 0.001, 64 (4 samples per GPU), 10, 1e-4 and 0.8, respectively.

\subsubsection{Hyperparameters for the model based on X3D}
On Kinetics400, we train our models on 32 GPUs.
The initial learning rate, batch size, total epochs, weight decay and dropout ratio are set to 0.4, 256 (8 samples per GPU), 256, 5e-5 and 0.5, respectively.
On Something-Something V1, the backbone networks are pretrained on Kinetics. 
We train our models on 32 GPUs, and these hyperparameters are set to 0.12, 256 (8 samples per GPU), 50, 4e-4 and 0.8, respectively.

\subsubsection{Hyperparameters for the model based on ViT}
The backbone networks are pretrained on ImageNet-21K~\cite{imagenet_cvpr09}.
In the training of ViT-based models, we do not use drop out, gradient clip and label-smoothing. 
The initial learning rate, batch size, total epochs, weight decay are set to 0.004, 64 (2 samples per GPU), 15, 1e-4, respectively. The learning rate for the PWTP module is 5 times larger than those for layers in ViT.

\section{Implementation details of Efficiency and Effectiveness of Dynamic Appearance}

These additional explanations are helpful for Section 4.2.4
from page 7-8 of the main paper.
In order to fairly evaluate different video representation methods~\cite{bilen2017action,fan2018end,ilg2017flownet,zach2007duality}, we reimplement these methods in the same experimental settings with the code provided by the original authors. In our reimplementation, Dynamic Image method generate one frame of representation for every 6 video frames, which is the same as our method. As for TVNet and TV-L1 Flow, we stack 5 frames of the estimated flow along the channel dimension to form an input frame with 10 channels, in total processing 6 RGB frames in a segment. All the models are pretrained on ImageNet-1K and then trained with the same hyperparameters.
\begin{table*}[ht]
\centering
 %%\vspace{0pt}
\small
% %%\vspace*{-0.3cm}

\setlength{\tabcolsep}{4pt}
\begin{tabular}{l c c c c}
\toprule
\multirow{2}*{Model} & \multicolumn{2}{c}{ Efficiency Metrics}  & \multirow{2}*{  SS V1} & \multirow{2}*{  UCF101}\\
\cmidrule{2-3}
 &  FLOPs &  \#Param. & & \\
\midrule
TSM R50~\cite{lin2019tsm} & 42.9G &  23.8 M  & 46.5 & 87.1 \\
TSM R50+Conv & 44.5G & 23.8M+2.6K & 46.3  & 87.0 \\
TSM R101~\cite{lin2019tsm} &  62.5G	&  42.8 M &	46.9 & - \\
\midrule
\textbf{DA TSM R50} & 43.1G & 23.8M+7.13K  & 48.7  & 89.7 \\
\bottomrule
\end{tabular}
\caption{\label{table: add_param}Comparisons with heavier models on Something-Something (SS) V1 and UCF101}
% %%\vspace*{-0.5cm} 
\end{table*}

In order to assess whether the performance gain of models that take dynamic appearances as inputs is or not due to the additional FLOPs and parameters, we compare our DA model with those models that have higher computational costs than ours. From the results shown in Fig.~\ref{table: add_param}, we can observe that our TSM R50 that takes dynamic appearances as input has even higher accuracy than TSM R101 (48.7\% \vs 46.9\%) on Something-Something, but the FLOPs and the number of parameters are much less than TSM R101, which demonstrate the performance gain of our model is more due to the efficient motion modeling in dynamic appearances. 

\section{Design of Multilayer Perceptron (MLP)}
\begin{figure}[ht]
\begin{center}
\resizebox{.38\textwidth}{!}{
\includegraphics{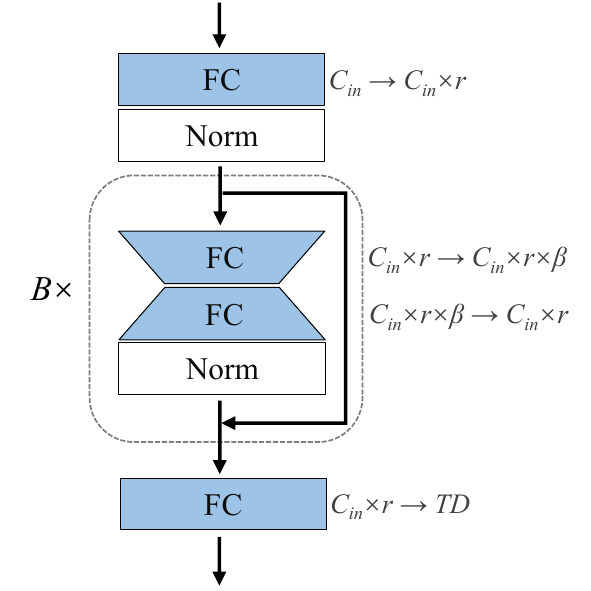} }
\end{center}
%%\vspace*{-0.2cm}
\caption{The scheme of MLP in the PWTP module}
\label{fig: mlp_PWTP}
\end{figure}
\noindent The MLP in the PWTP module is used to generate the input-specific hyperplanes $\mathbf{A}$ in Equation~(4) from Section 3.1, page 5 of the main paper. In Figure~\ref{fig: mlp_PWTP}, we present the scheme of the MLP, where the feature numbers of the fully-connected (FC) layers are specified on their right side. The first FC expends the number of features with a factor of $r$. The MLP contains $B$ bottleneck blocks. Each bottleneck block comprises a normalization layer and two FC layers, where the number of features is first reduced by a factor of $\beta$ and then recovered to its original size. For the normalization, we use Batch Normalization, which behaves more stably than Layer Normalization in our study. In Equation ~(4) from the main paper, the number of input channels for the MLP is $C_{in}=\frac{T \times (T-1)}{2}$, while the number of output channels is $TD$.
\begin{table*}[ht]
    \centering
    \small

    \setlength{\tabcolsep}{2pt}
    \begin{tabular}{c c c c c c c}
    \toprule
     \multirow{2}*{Config} & Bottleneck    &   Number of  &  Expansion  & \multirow{2}*{ENoPR ($\widehat{\mathcal{L}}^{1}$)}& \multirow{2}*{FLOPs} & \multirow{2}*{Params}\\
     & ratio ($\beta$)   &   blocks ($B$) & ratio ($r$) & &  & \\
    \midrule
    \#1  & 1/4 & 1 & 4 & 0.26& 45.3M & 16.9k\\
    \#2  & 1/4 & 1 & 1 & 0.27& 38.0M & 7.4k\\
    \#3  & 1/4 & 2 & 1 & 0.28& 38.3M & 7.9k\\
    \#4  & 1/4 & 2 & 4 & 0.26& 50.2M & 23.3k\\
    \#5  & 1 & 1 & 4 & 0.28 & 59.8M & 35.5k\\
    \#6  & 1/4 & 0 & 4 & 0.30 & 40.0M & 10.0k\\
    \#7  & 1/4 & 1 & 2 & 0.27 & 39.7M & 9.7k\\
    \bottomrule
    \end{tabular}
        \caption{Various configurations of the MLP (Lower ENoPR ($\widehat{\mathcal{L}}^{1}$) means higher capacity for representation). We report the computational cost (FLOPs) of the PWTP module}
    \label{tab: config_mlp}

\end{table*}
Here, we treat PWTP as an individual optimization problem and evaluate its representation capacity by ENoPR $\widehat{\mathcal{L}}^{1}(\Theta^1)$ on Mini-Kinetics. The specifications of the MLP's various configurations are listed in Table~\ref{tab: config_mlp}, in which the PWTP modules are trained for 10 epochs with AdamW optimizer~\cite{loshchilov2018decoupled}. We can see that aside from Config~\#6, which does not use bottleneck blocks, the rest of the configurations have similar ENoPR ($\widehat{\mathcal{L}}^{1}$). To balance the capacity/speed trade-off, we use Config~\#7 as the default MLP configuration in our experiments.

\section{Scale Scheduler for Joint Training}
\begin{figure}[ht]
    \centering
    % %%\vspace*{-0.2cm}
    \resizebox{.45\textwidth}{!}{
    \includegraphics{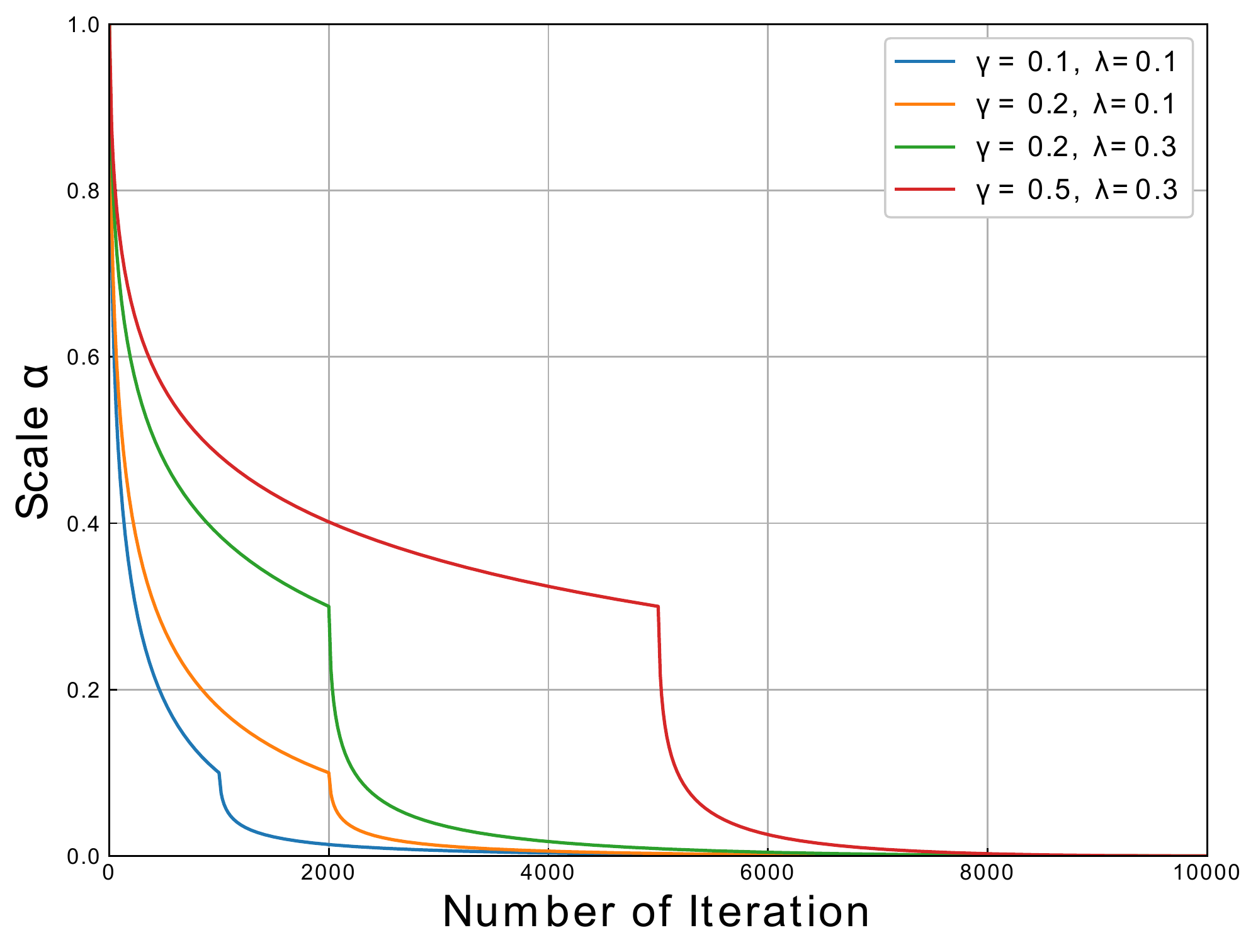}
    }
    \caption{Example graphs of the scale schedule function with different $\gamma$ and $\lambda$}
    \label{fig:plot_cosine_mtl}
    %%\vspace*{-0.5cm}
\end{figure}

When using joint training to resolve the multi-objective optimization problem defined in~Equation~(9) from the paper, we also consider employing a scale scheduler to produce a dynamic scale $\alpha$, weighting the contribution of the two components $\widehat{\mathcal{L}}^{1}(\Theta^1)$ and $\widehat{\mathcal{L}}^{2}(\Theta^1,\Theta^2)$. The scale scheduler is a hybrid function that initially sets $\alpha$ to a large value and gradually decreases it to a minimum value, simulating the cosine annealing~\cite{loshchilov2016sgdr}:
\begin{equation}
% %%\vspace*{-0.3cm}
\label{eq:cosine_decay_mtl}
\begin{aligned}
\alpha = 
\begin{cases}
\frac{1}{2}\big(1+\cos(  \pi \log_{{ \big[(\gamma M)^{\tfrac{\pi}{\cos^{-1}(2\lambda -1)}} \big ] }}m)\big), & m < \gamma M\\
\frac{1}{2} \lambda \big(1+\cos(\pi {\log_{\big[(1- \gamma)M\big]}}m) \big), & m \geq  \gamma M
\end{cases}
\end{aligned}
\end{equation}
where $m$ denotes the current iteration, $M$ represents the total number of training iterations, while $\gamma \geq 1$ and  $\lambda \leq 1 $ are two additional parameters used to control the graph of the hybrid function. 
Some example graphs of the hybrid function are displayed in Figure~\ref{fig:plot_cosine_mtl}.
The scale scheduler is established under the realistic assumption that in the early training period, the feature representation learning task $\widehat{\mathcal{L}}^1(\Theta^1)$ has higher priority than the recognition task $\widehat{\mathcal{L}}^{2}(\Theta^1,\Theta^2)$. In the later training period, the primary action recognition task should be the main focus, and the optimization objective of the feature representation learning task should be assigned a very low weight.
\begin{table}[ht]
% %%\vspace*{-0.2cm}
    \centering
    \setlength{\tabcolsep}{7pt}
    \small
    \begin{tabular}{l c c}
    \toprule
     Scale Scheduler & ENoPR ($\widehat{\mathcal{L}}^{1}$) & Accuracy (\%) \\
    \midrule
    $\gamma=0.1, \lambda=0.1$& 0.57 & 40.7  \\
    $\gamma=0.2, \lambda=0.1$& 0.53 & 40.6 \\
    $\gamma=0.2, \lambda=0.3$& 0.56 & 40.7\\
    $\gamma=0.5, \lambda=0.3$& 0.55 & 40.4\\
    \bottomrule
    \end{tabular}
        \caption{\label{tab: scale_scheduler}Joint training with the Scale Scheduler on Something Something V1}

\end{table}
We use the small network X3D-XS~\cite{feichtenhofer2020x3d} as the backbone network processing $S=4$ segments. 
The results of using the scale scheduler are shown in 
Table~\ref{tab: scale_scheduler}.
The scale scheduler works better than the fixed scale method. 
Although the scale scheduler introduces two additional parameters $\lambda$ and $\gamma$,  its performance is not that sensitive to changes in these parameters, as can be observed from the results in Table~\ref{tab: scale_scheduler}.

\section{More visualization analysis results}

\subsubsection{Dynamic Appearance of longer sequences}
\begin{figure*}[!t]
\begin{center}
\resizebox{1.\textwidth}{!}{
\includegraphics{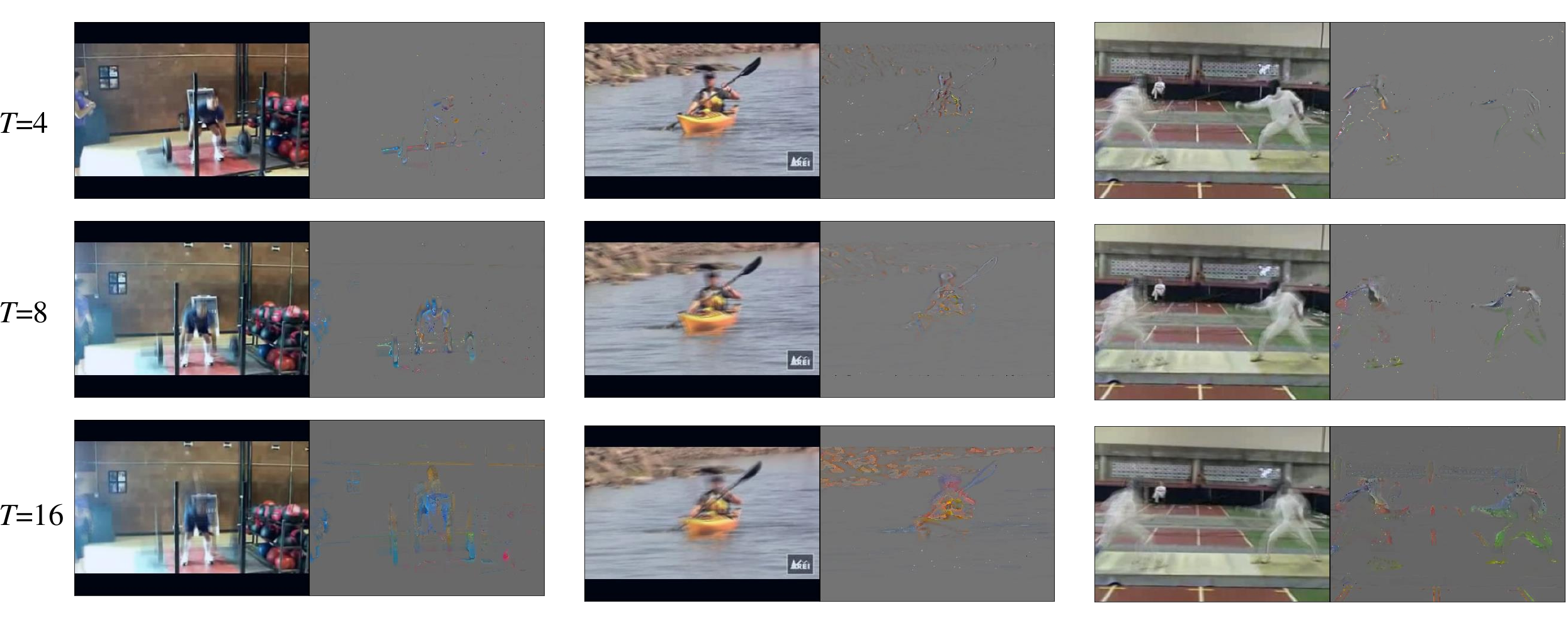} }
\end{center}
\caption{Example videos of different lengths ($T$) visualized as their frame averaging and corresponding dynamic appearances.}
\label{fig: example_diff_len}
\end{figure*}

\noindent We attempt to extract the dynamic appearances from video frame sequences of different lengths ($T$). Some visualization examples are presented in Figure~\ref{fig: example_diff_len}. We can observe that the dynamic appearances generated by longer frame sequences contain additional visual information characterizing the movement. According to the results in Table~\ref{tab: diff_len},  the dynamic appearances extracted from longer frame sequences have higher accuracy. We do not experiment with longer frame sequences on the Something-Something dataset because many videos in that dataset contain fewer than 50 frames, which is not enough to form a valid input when X3D-XS processes $S=4$ segments. 
\begin{table}[!t]
    \centering
    \small
    \begin{tabular}{l c c}
    \toprule
    $T$ &  ENoPR ($\widehat{\mathcal{L}}^{1}$) & Accuracy\\
    \midrule
     4  & 0.26  & 56.2  \\
     8 & 0.41  & 56.9  \\
     16 & 053  & 58.2  \\
    \bottomrule
    \end{tabular}
    \caption{Evaluating dynamic appearances generated with various temporal lengths $T$ on Mini-Kinetics.}
    \label{tab: diff_len}
\end{table}

\subsubsection{More dynamic appearance examples}
\begin{figure*}[!t]
\begin{center}
\resizebox{1.\textwidth}{!}{
\includegraphics{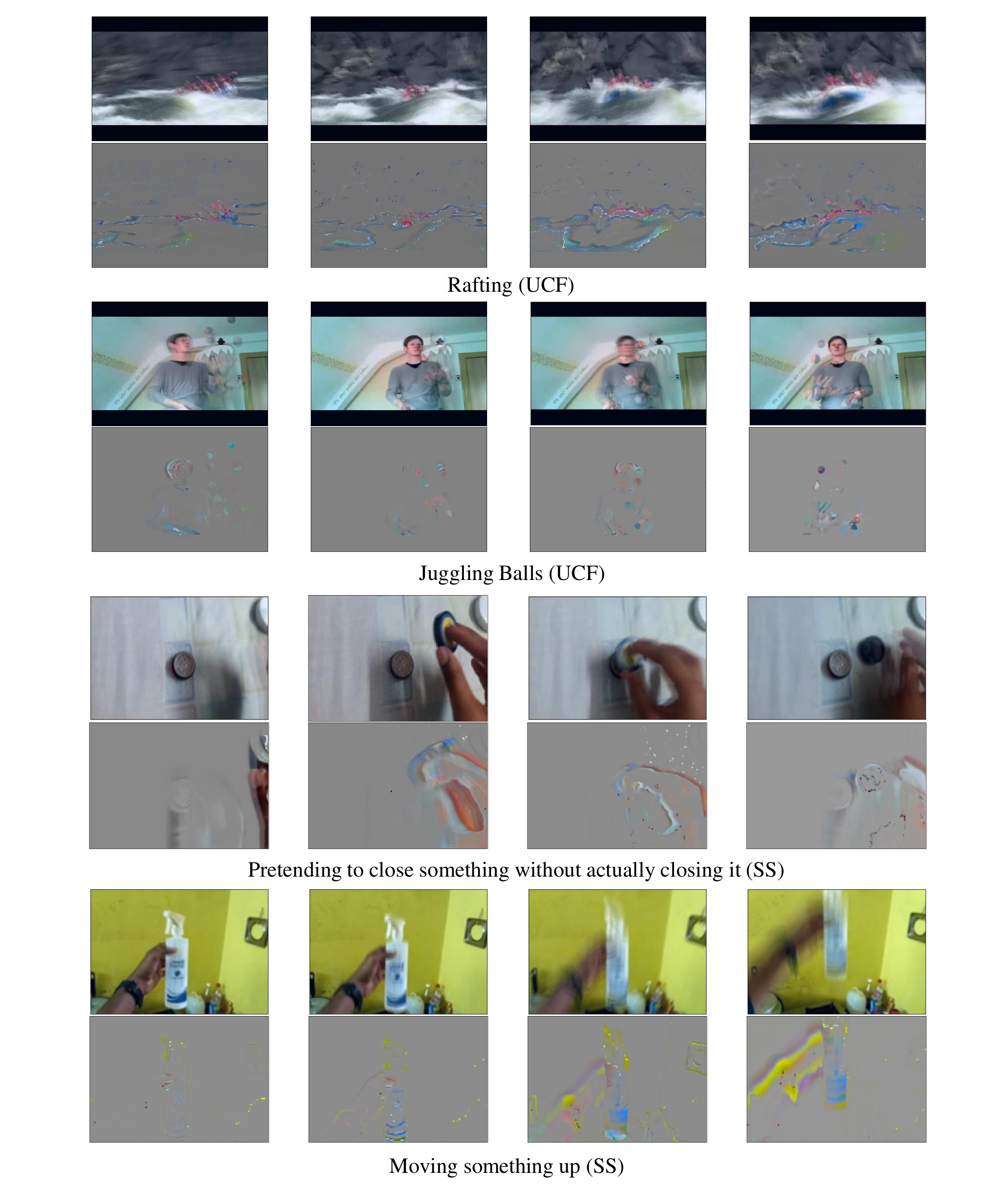} }
\end{center}
\caption{Example videos of 4 segments visualized as their frame averaging and corresponding dynamic appearances.}
\label{fig: more_example1}
\end{figure*}

\begin{figure*}[!t]
\begin{center}
\resizebox{1.\textwidth}{!}{
\includegraphics{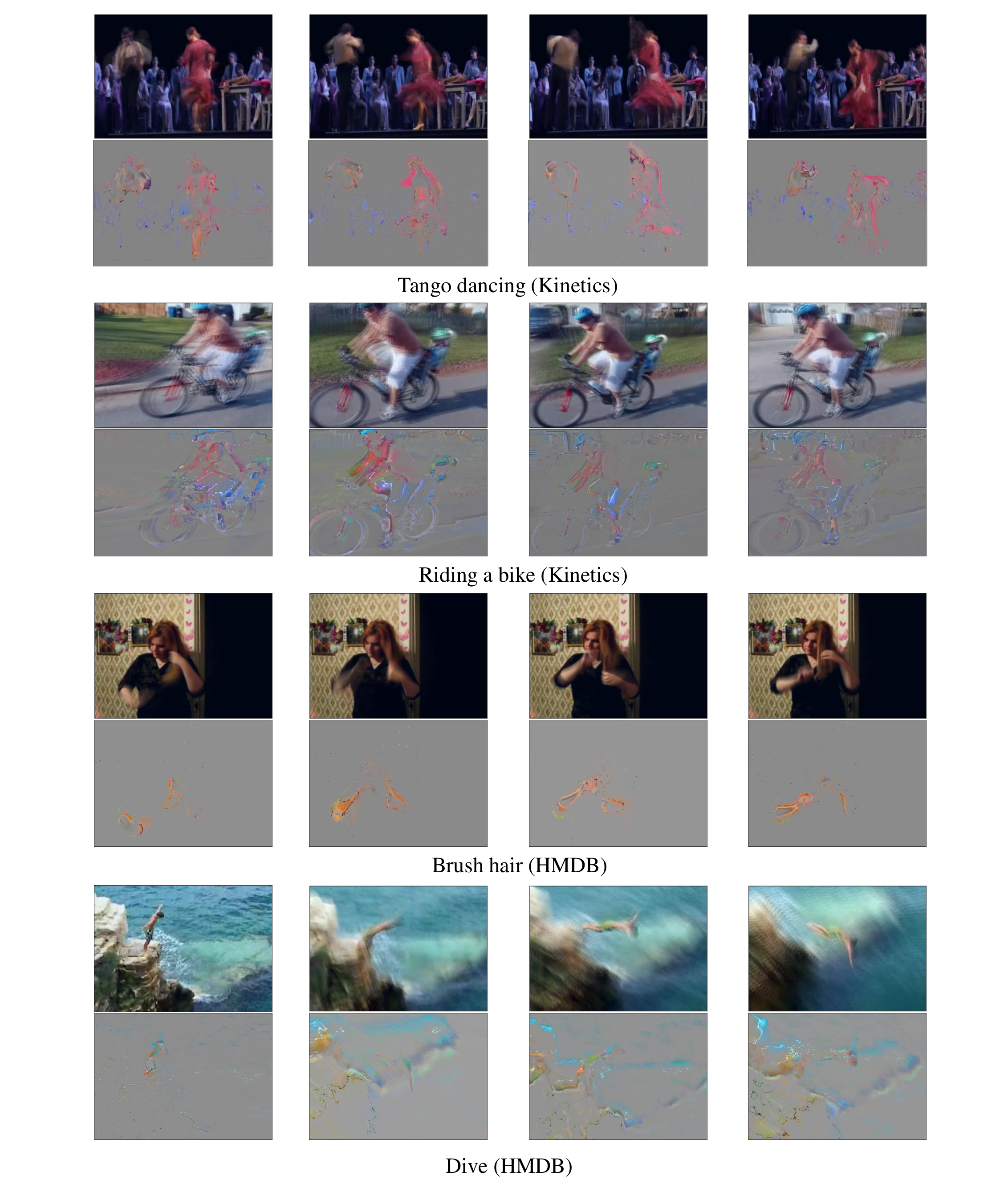} }
\end{center}
\caption{Example videos of 4 segments visualized as their frame averaging and corresponding dynamic appearances.}
\label{fig: more_example2}
\end{figure*}

We provide more visualization examples in Figures~\ref{fig: more_example1} and~\ref{fig: more_example2}, where we provide the averaging of sets of frames from video-clips and underneath their corresponding Dynamic Appearances. From these results, containing a wide range of videos, with 2 examples from each of UCF, Something Something V1, Kinetics and HMDB database, we can observe that the dynamic appearance extracted by PWTP retains the essence of movement for a wide range of actions.

\end{document}